%% file: blinq_cvpr.tex
\documentclass[10pt,twocolumn,letterpaper]{article}
\usepackage[usenames,dvipsnames]{xcolor}

\usepackage{cvpr}
\usepackage{times}
\usepackage{epsfig}
\usepackage{graphicx}
\usepackage{amsmath}
\usepackage{amssymb}
\usepackage{multirow}
\usepackage{rotating}
\usepackage{commath}
\usepackage{tikz,pgfplots}
\usepackage{arydshln}
\usepackage{wrapfig}

\newlength{\tikzpictureheight}
\newlength{\tikzpicturewidth}
\usepackage[pagebackref=true,breaklinks=true,letterpaper=true,colorlinks,bookmarks=false]{hyperref}

\cvprfinalcopy % *** Uncomment this line for the final submission

\def\cvprPaperID{116} % *** Enter the CVPR Paper ID here

\ifcvprfinal\fi
\begin{document}

\newcommand{\todo}[1]{\textcolor{red}{[#1]}}

%%%%%%%%% TITLE
\title{Some like it hot - visual guidance for preference prediction}

%\author{Rasmus Rothe\\
%Computer Vision Lab\\
%D-ITET, ETH Zurich\\
%{\tt\small rrothe@vision.ee.ethz.ch}
%\and
%Radu Timofte\\
%Computer Vision Lab\\
%D-ITET, ETH Zurich\\
%{\tt\small radu.timofte@vision.ee.ethz.ch}
%\and
%Luc Van Gool\\
%ESAT, KU Leuven\\
%D-ITET, ETH Zurich\\
%{\tt\small vangool@vision.ee.ethz.ch}
%}

\author{Rasmus Rothe\\
CVL, ETH Zurich\\
{\tt\small rrothe@vision.ee.ethz.ch}
\and
Radu Timofte\\
CVL, ETH Zurich\\
{\tt\small radu.timofte@vision.ee.ethz.ch}
\and
Luc Van Gool\\
KU Leuven, ETH Zurich\\
{\tt\small vangool@vision.ee.ethz.ch}
}

\maketitle
%\thispagestyle{empty}

%%%%%%%%% ABSTRACT
\begin{abstract}
\vspace*{-0.4cm}
For people first impressions of someone are of determining importance. 
They are hard to alter through further information.
This begs the question if a computer can reach the same judgement.
Earlier research has already pointed out that age, gender,
and average attractiveness can be estimated with reasonable precision. We improve the 
state-of-the-art, but also predict - based on someone's known preferences - how much 
that particular person is attracted to a novel face. Our computational pipeline comprises
a face detector, convolutional neural networks for the extraction of deep features, 
standard support vector regression for gender, age and facial beauty, and - as the main 
novelties - visual regularized collaborative filtering to infer inter-person preferences as well as a novel regression technique for handling visual queries without rating history.
We validate the method using a very large dataset from a dating site as well as images from celebrities.
Our experiments yield convincing results, \ie we predict $76\%$ of the ratings correctly solely based on an image, and reveal some sociologically relevant conclusions.
We also validate our collaborative filtering solution on the standard MovieLens rating dataset, augmented with movie posters, to predict an individual’s movie rating. We demonstrate our algorithms on \url{howhot.io} which went viral around the Internet with more than 50 million pictures evaluated in the first month.
\end{abstract}

%%%%%%%%% BODY TEXT
\vspace*{-0.5cm}
\section{Introduction}
\label{sec:introduction}
\vspace*{-0.1cm}
`First impressions count' the saying goes. Indeed, psychology confirms that it only takes 
0.1 seconds for us to get a first impression of someone~\cite{Willis-PS-2006}, with the  
face being the dominant cue. Factors that are relevant for survival seem the ones evolution 
makes us pick up the fastest. These include age, gender, and attractiveness. We will call 
those quantifiable properties, such as age and gender, `demographics'.
\\
An everyday demonstration is that people on dating sites often base their decisions mainly 
on profile images rather than textual descriptions of interests or occupation. Our goal 
is to let a computer predict someone's preferences, from single facial photos (in the wild). 
In particular, we try to predict how much a previously unseen face would be attractive 
for a particular person who has already indicated preferences for people in the
system.  

\begin{figure}[t!]
\begin{center}
\vspace*{-0.2cm}
\includegraphics[width=0.8\linewidth]{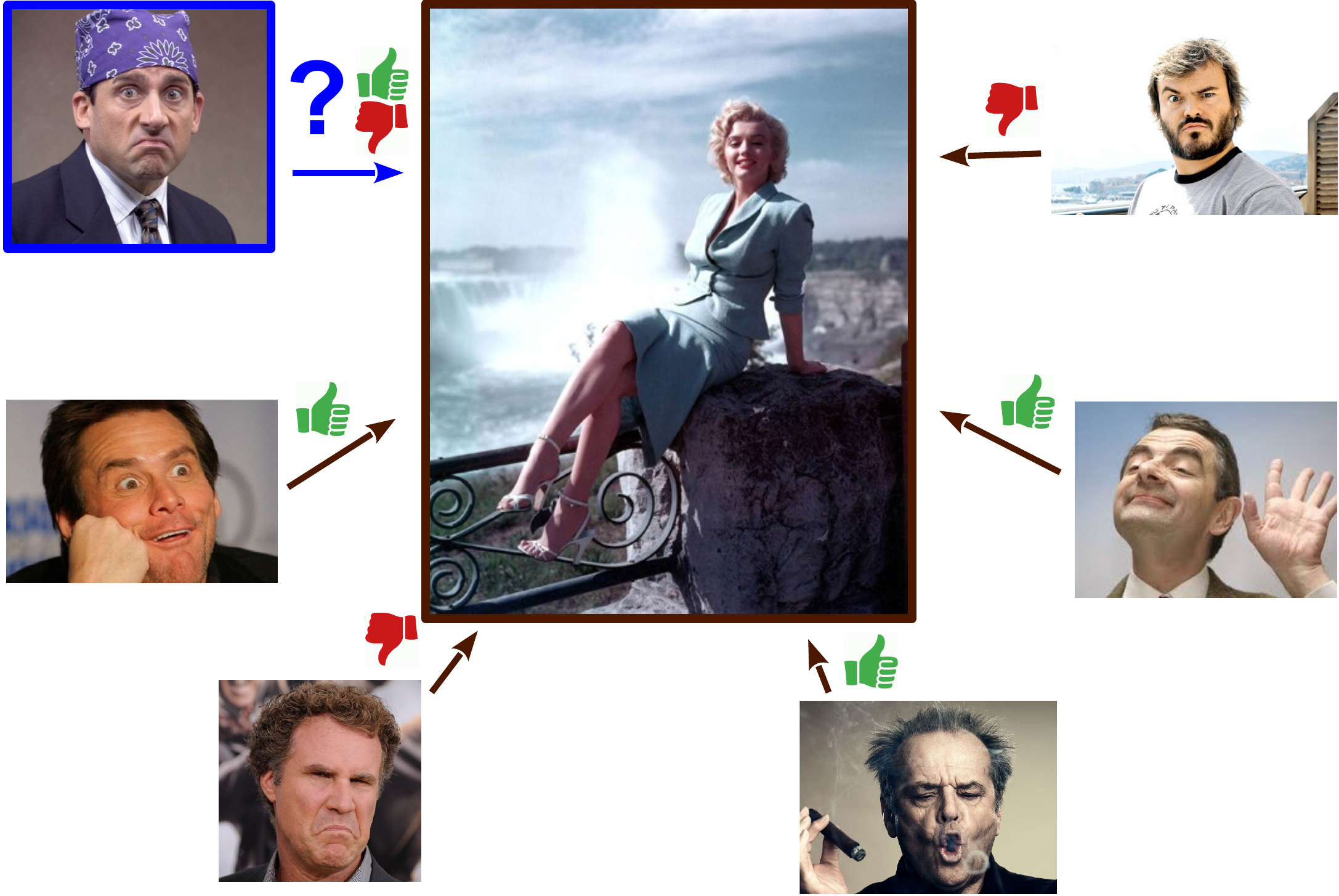}
\end{center}
\vspace*{-0.4cm}
\caption{Can we infer preferences from a single image?}
\vspace*{-0.5cm}
\label{fig:teaser}
\end{figure}

Our main benchmark is a large dataset of more than 13,000 user profiles from a dating site. We have access to their age and gender, as well as more than 17 million `hot or not' ratings by some users about some other users (their profile photo). The ratings are very sparse when compared to their potential number. For people who have given ratings, we want to predict new ratings for other people in and outside the dataset. 

The visual information, here the profile image, presumably containing a face, is the main basis for any user-to-user rating. Therefore, we employ a face detector and crop the best detected face and its surrounding context (corresponding to body and posture) from which we extract deep features by means of a (fine-tuned) convolutional neural network. In order to make sure that these features are appropriate for the main task - automated attractiveness rating - we first test our features on age, gender, and facial beauty estimation for which previous methods and standard datasets exist. We obtain state-of-the-art results.

For predicting preferences for users with known ratings for a subset of others in the dataset, collaborative filtering is known to provide top results, \ie for movie~\cite{koren2009matrix} or advertisement suggestions~\cite{robinson1999automated}. We adapt this framework to take account of visual information, however. As our experiments will show, adding visual information improves the prediction, especially in cases with few ratings per user. In case of a new face, not part of the dataset and thus without a history of preferences, we propose to regress the input image to the latent space of the known users. By doing so, we alleviate the need for past ratings for the query and solely rely on the query image.

The same technique can be applied to different visuals-enhanced tasks, such as rating prediction of movies, songs, shopping items, in combination with a relevant image (\eg movie poster, CD cover, image of the item). We test on the MovieLens dataset augmented with poster images for each movie, a rather weak information, to demonstrate the wider applicability of our approach.

We demonstrate our algorithms on \url{howhot.io}, a website where people can upload a photo of their face and an algorithm will then estimate the age, gender and facial attractiveness of the person.
%The website went viral around the Internet with more than 50 million pictures evaluated in the first month.

The main contributions of our work are:
\begin{itemize}
\vspace*{-0.2cm}
\itemsep0em
\item an extensive study on the inference of information from profile images using the largest dating dataset thus far
\vspace*{-0.1cm}
\item a novel collaborative filtering approach that includes visual information for rating/preference prediction
\vspace*{-0.1cm}
\item a novel regression technique for handling visual queries without rating history which prior work cannot cope with
\end{itemize}

%The paper is organized as follows. Section~\ref{sec:relatedwork} reviews the related work.
%Section~\ref{sec:demographics} describes our approach on the estimation of age, gender and facial beauty (hotness) by use of convolutional neural networks to extract deep features and of support vectors regression. 
%Section~\ref{sec:method} presents our proposed method to predict ratings/preferences. We start from the standard collaborative filtering framework and propose visual regularization, and tackle the rating of history-free faces by means of piece-wise linear regression of the missing information. 
%Section~\ref{sec:experiments} describes the main experimental results on the dating and MovieLens datasets. 
%Section~\ref{sec:conclusions} concludes the paper.
\vspace*{-0.5cm}
\section{Related Work}
\label{sec:relatedwork}
\vspace*{-0.15cm}

The focus of our paper is to infer as much information as possible from a single image and to predict subjective preferences based on an image query with possibly a prior rating history.
Next we review related works.

\noindent{\bf Image features.} 
Instead of handcrafted features like SIFT, HoG, or Gabor filters, we use learned features obtained using neural networks~\cite{girshick2014rcnn,krizhevsky2012imagenet,simonyan2014very}. The latter have shown impressive performance in recent years. Such features have already been used for age and gender estimation in~\cite{Rothe-ICCVW-2015,wang2015deeply}. 
%We use the recently proposed convolutional neural networks (CNN) features from the classification literature~\cite{simonyan2014very}. 

%An image is either used in its raw form (pixel values), or after undergoing a feature extraction process. We distinguish handcrafted features such as SIFT~\cite{lowe2004distinctive}, HoG~\cite{dalal2005histograms}, Gabor filters~\cite{jain1990unsupervised}, or biologically inspired features (BIF)~\cite{Han-ICB-2013} from learned features such as the ones obtained using neural networks~\cite{girshick2014rcnn,krizhevsky2012imagenet,simonyan2014very}. The latter have shown impressive performance in recent years. Neural networks or similar features have been used for face demographics estimation in~\cite{wang2015deeply}. We use the recently proposed convolutional neural networks (CNN) features from the classification literature~\cite{simonyan2014very}. 

\noindent{\bf Demographics estimation.} Multiple demographic properties such as age, gender, and ethnicity have been extracted from faces. A survey on age prediction is provided by Fu~\etal~\cite{fu2010age} and on gender classification by Ng~\etal~\cite{ng2012recognizing}.
Kumar~\etal~\cite{Kumar-ICCV-2009} investigate image `attribute' classifiers in the context of face verification. 
Some approaches need face shape models or facial landmarks~\cite{Han-ICB-2013,Jaimes-CVIU-2007}, others are meant to work in the wild~\cite{chang2011OHRank,chen2013cumulative,Rothe-ICCVW-2015,wang2015deeply} but still assume face localization.
Generally, the former approaches reach better performance as they use additional information. The errors in model fitting or landmark localization are critical. Moreover, they require supervised training, detailed annotated datasets, and higher computation times. 
On top of the extracted image features a machine learning approach such as SVM~\cite{Vapnik-1998} is employed to learn a demographics prediction model which is then applied to new queries. 
%Our demographics estimation uses Support Vector Regression~\cite{CC01a}.

\noindent{\bf Subjective property estimation.} 
While age and gender correspond to objective criteria, predicting the attractiveness of a face is more subjective.
Nonetheless, facial beauty~\cite{altwaijry2013relative,eisenthal2006facial,gray2010predicting,mu2013computational,yan2014cost} can still be quantified by averaging the ratings by a large group. Benchmarks and corresponding estimation methods have been proposed. 
In the subjective direction, Dhar~\etal~\cite{Dhar-CVPR-2011} demonstrate the aesthetic quality estimation and predict what they call `interestingness' of an image, while Marchesotti~\etal~\cite{Marchesotti-IJCV-2014} discover visual attributes (including subjective ones) to then to use them for prediction. Also, recently Kiapour~\etal~\cite{Kiapour-ECCV-2014} inferred complex fashion styles from images.
Generally, the features and methods used for age and gender can be adapted to subjective property estimation, and we do the same in this paper. From the literature we can observe three trends:
(i) besides Whitehill and Movellan~\cite{whitehill2008personalized}, most papers focus on predicting facial beauty averaged across all ratings, whereas we aim at predicting the rating by a specific person; 
(ii) as pointed out in the study by Laurentini and Bottino~\cite{laurentini2014computer} usually small datasets are used, sometimes with less than 100 images and with only very few ratings per image -- our dataset contains more than 13,000 images with more than 17 million ratings;
(iii) most datasets are taken in a constrained environment showing aligned faces. In contrast, our photos contain in many cases also parts of the body and some context in the background. Thus, we focus not just on facial beauty but on general attractiveness of the person -- referred to as hotness in the following.

\noindent{\bf Preferences/ratings prediction.} The Internet brought an explosion of choices. Often, it is difficult to pick suitable songs to hear, books to read, movies to watch, or - in the context of dating sites - persons to contact. Among the best predictors of interest are collaborative filtering approaches that use the knowledge of the crowd, \ie the known preferences/ratings of other subjects~\cite{breese1998empirical,sarwar2001item}. The more prior ratings there are, the more accurate the predictions become. Shi~\etal~\cite{shi2014collaborative} survey the collaborative filtering literature.
Matrix factorization lies at the basis of many top collaborative filtering methods~\cite{koren2009matrix,lee2001algorithms}. Given the importance of the visual information in many applications, we derive a matrix factorization formulation regularized by the image information. While others~\cite{shi2014collaborative} proposed various regularizations, we are the first to prove that visual guidance helps preference prediction. Moreover, we propose to regress queries without rating history to a latent space derived through matrix factorization for the known subjects and ratings.

\begin{figure*}[th!]
\begin{center}
\includegraphics[width=\linewidth]{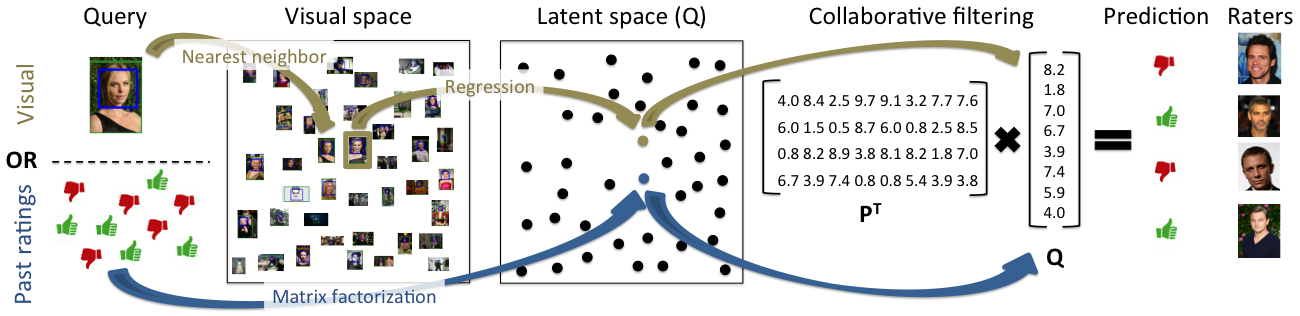}
\end{center}
\vspace*{-0.4cm}
\caption{Preferences prediction scheme. For a visual query without past ratings we first regress to the latent Q space (obtained through matrix factorization) to then obtain the collaborative filtering prediction as in the case for the queries with known past ratings and Q factor.}
\vspace*{-0.5cm}
\label{fig:pipeline}
\end{figure*}

\noindent{\bf Social networks.} The expansion of the internet and the advance of smartphones boosted the (online) social networks worldwide. Networks such as Facebook facilitate interaction, sharing, and display of information and preferences among individuals. Yet, time is precious and hence efforts are made to develop filters and ranking tools to assist users. A recent study by Youyou~\etal~\cite{youyou2015computer} shows that accurate predictions about the personality of a user can be made using her/his `likes'. Contents and ads can then be personalized and this is extremely important for social networks and search engines such as Google~\cite{Sugiyama-WWW-2004}. This paper focuses on dating sites and the prediction of attractiveness ratings. Most such works~\cite{brozovsky2007recommender,krzywicki2010interaction} rely on past ratings and cannot cope when there are none or few.

\vspace*{-0.2cm}
\section{Visual features}
\label{sec:demographics}
\vspace*{-0.1cm}

Razavian~\etal~\cite{razavian2014cnn} showed that features extracted from convolutional neural networks (CNN) are very powerful generic descriptors. Inspired by that, for all our experiments we use the VGG-16~\cite{simonyan2014very} features which are pre-trained on a large ImageNet object dataset and result in a descriptor of length 4,096. We use the implementation by Vedaldi and Lenc~\cite{vedaldi2014matconvnet}. We reduce the dimensionality using PCA to keep $\sim99\%$ of the energy. Before we use these feature to predict attractiveness to a particular user, we first confirm that the extracted visual features are powerful enough to capture minor facial differences by predicting age and gender.

We perform reference experiments on a widely used dataset for age prediction, the MORPH 2 database~\cite{ricanek2006morph}. We also test gender estimation on the same MORPH 2 dataset. Unfortunately, besides the dataset provided by Gray~\etal~\cite{gray2010predicting} -- to the best of our knowledge -- there are no other publicly available large datasets on averaged facial beauty. As shown next our features achieve state-of-the-art performance for age, gender, and facial beauty prediction. We believe that this good performance is mostly due to the depth of the model with 16 layers, compared with previous state-of-the-art using only 6 layers~\cite{wang2015deeply}. 

\subsection{Predicting age and gender}
Our experiments are conducted on a publicly available dataset, the MORPH 2 database~\cite{ricanek2006morph}. 
We adopt the experimental setup of~\cite{chang2011OHRank,chen2013cumulative,guo2008image,wang2015deeply}, where a set of 5,475 individuals is used whose age ranges from 16 to 77. The dataset is randomly divided into 80\% for training and 20\% for testing. Following the procedure described in~\cite{girshick2014rcnn}, our CNN features are fine-tuned on the training set.

The age is regressed using Support Vector Regression (SVR)~\cite{CC01a} with an RBF kernel and its parameters are set by cross-validation on a subset of the training data. We report the performance in terms of mean absolute error (MAE) between the estimated and the ground truth age.

\begin{table}[h]
\scriptsize
\begin{center}
\begin{tabular}{cc}
Method                                & MORPH 2~\cite{ricanek2006morph} \\ % &FG-NET~\cite{fgnet2015}
\hline
%Human workers~\cite{han2014demopgrahic}& 6.30 & 4.70  \\
%DIF~\cite{han2014demopgrahic}          & 3.80 & 4.80  \\
AGES~\cite{geng2007automatic}           & 8.83 \\ %& 6.77
MTWGP~\cite{zhang2010multi}             & 6.28 \\ %& 4.83
CA-SVR~\cite{chen2013cumulative}        & 5.88\\ %& 4.67
SVR~\cite{guo2008image}                & 5.77\\ % & 5.66
OHRank~\cite{chang2011OHRank}           & 5.69\\ %& 4.85
DLA~\cite{wang2015deeply}               & 4.77\\ %& {\bf 4.26}
Proposed Method                       & {\bf 3.45} \\ % & 5.01$^*$ 
\hline
\end{tabular}
\end{center}
\vspace*{-0.2cm}
\caption{Age estimation performance in terms of mean absolute error (MAE) on the MORPH 2 dataset. We improve the state-of-the-art results by more than 1 year.}
\vspace*{-0.25cm}
\label{tab:ageprediction}
\end{table}

\begin{figure}[th!]
\vspace*{-0.2cm}
\begin{center}
\setlength{\tabcolsep}{1pt}
\begin{tabular}{cc}
Age in MORPH 2, men&Beauty in Gray \\
\includegraphics[height=1cm]{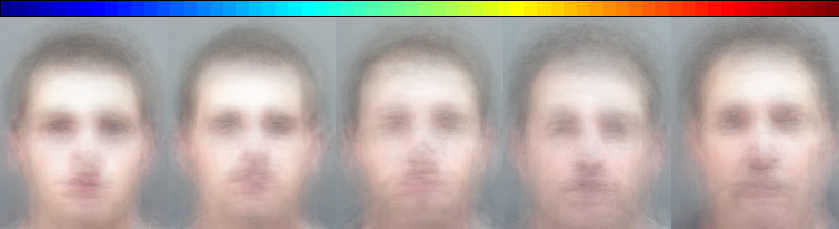}&
\includegraphics[height=1cm]{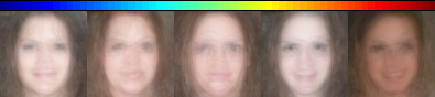}\\
\end{tabular}
\end{center}
\vspace*{-0.3cm}
\caption{Average faces for 5 clusters based on age or beauty, resp. Average beauty is less meaningful, suggesting personalized prediction.}
\vspace*{-0.3cm}
\label{fig:meanface}
\end{figure}

As shown in Table~\ref{tab:ageprediction}, we achieve state-of-the-art performance on the MORPH 2 dataset ($3.45$ years MAE) by reducing the error of the currently best result ($4.77$ MAE reported by~\cite{wang2015deeply}) with more than a full year. For gender prediction on the MORPH 2 dataset we keep the same partition as for age and achieve $96.26\%$ accuracy, which, despite the small training set, is on par with other results in the literature~\cite{guo2011simultaneous,guo2013joint}. Fig.~\ref{fig:agegenderbeautyexamples} shows several good and erroneous predictions of our method on the MORPH 2 dataset. 
Fig.~\ref{fig:meanface} shows averages of faces ranked according to age on MORPH 2 and beauty on Gray, resp. 
On our more challenging Hot-or-Not dataset (Section \ref{sec:dataset}) we achieve $3.61$ MAE for age and $88.96\%$ accuracy for gender prediction. 

\subsection{Predicting facial beauty}
Following a similar procedure as for age prediction, we test our features on the dataset introduced by Gray~\etal~\cite{gray2010predicting}. The Gray dataset contains 2056 images with female faces collected from a popular social/dating website\footnote{http://hotornot.com}. The facial beauty was rated by 30 subjects and the ratings were then normalized as described in~\cite{gray2010predicting}. The dataset is split into 1028 images for training and 1028 for testing. We report the average performance across exactly the same 5 splits from the reference paper in terms of Pearson's correlation coefficient, the metric from the original paper. Also, we report performance with and without face alignment using the same alignment algorithm of Huang~\etal~\cite{huang2007unsupervised}.

\begin{table}[h!]
\scriptsize
\begin{center}
\begin{tabular}{ccc}
\multirow{2}{*}{Method}                      & Correlation  & Correlation \\
                                             & w/o alignment& w/ alignment\\
\hline
Eigenface                                    & 0.134        & 0.180       \\
Single Layer Model~\cite{gray2010predicting} & 0.403        & 0.417       \\
Two Layer Model~\cite{gray2010predicting}    & 0.405        & 0.438       \\
Multiscale Model~\cite{gray2010predicting}   & 0.425        & 0.458       \\
Proposed Method                              & {\bf 0.470}  & {\bf 0.478} \\
\hline
\end{tabular}
\end{center}
\vspace*{-0.3cm}
\caption{Facial beauty estimation performance on Gray dataset with and without face alignment in terms of correlation.}
\vspace*{-0.2cm}
\label{tab:hotness}
\end{table}

\begin{figure*}[ht!]
\begin{center}
\setlength{\tabcolsep}{1pt}
\footnotesize
\begin{tabular}{lcccccc@{\hskip 0.25cm}cccc@{\hskip 0.25cm}cccccc}
&\multicolumn{6}{c}{MORPH 2 (age)}&\multicolumn{4}{c}{MORPH 2 (gender)}    &\multicolumn{6}{c}{Gray (facial beauty)}  \\
&
\includegraphics[height=1.0cm]{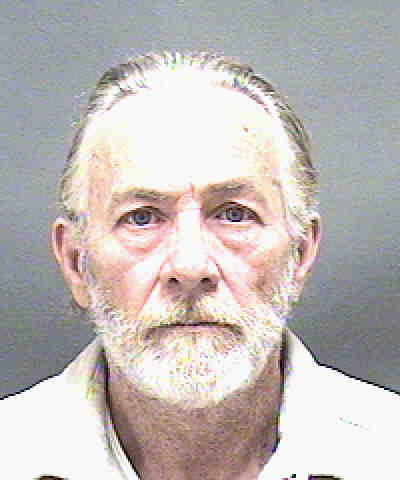}&
\includegraphics[height=1.0cm]{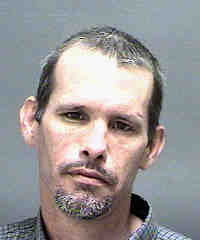}&
\includegraphics[height=1.0cm]{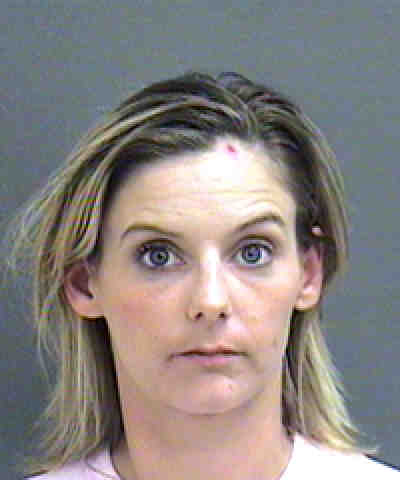}&
\includegraphics[height=1.0cm]{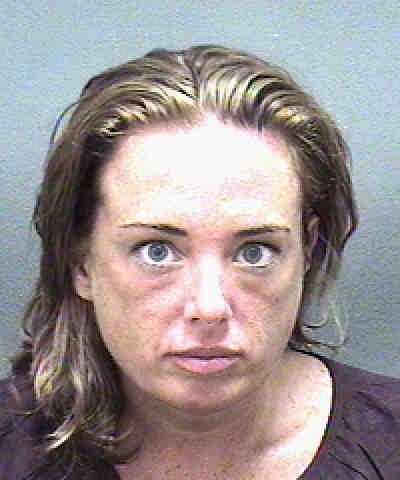}&
\includegraphics[height=1.0cm]{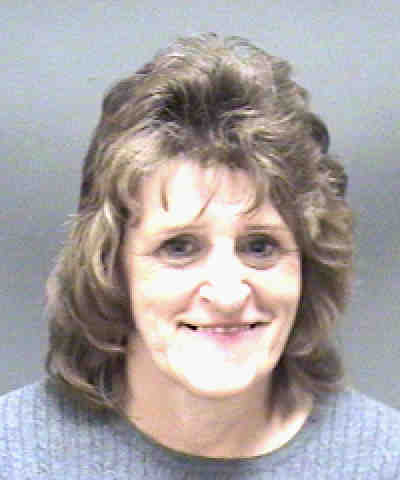}&
\includegraphics[height=1.0cm]{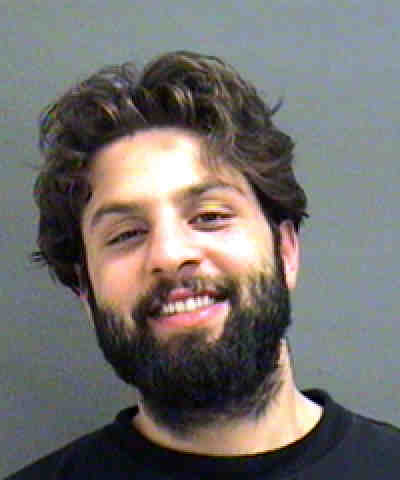}&
\includegraphics[height=1.0cm]{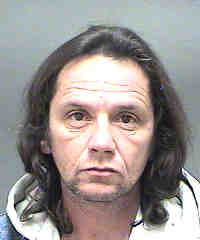}&
\includegraphics[height=1.0cm]{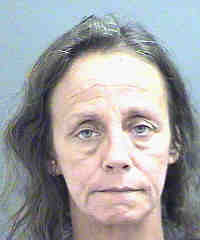}&
\includegraphics[height=1.0cm]{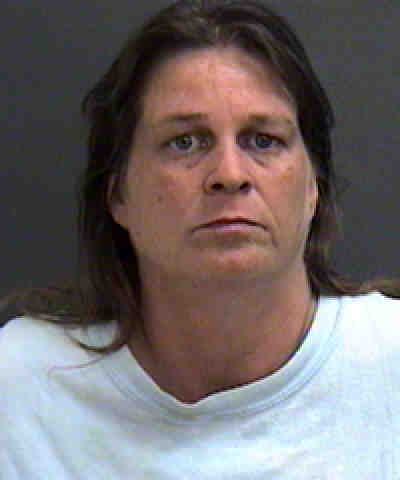}&
\includegraphics[height=1.0cm]{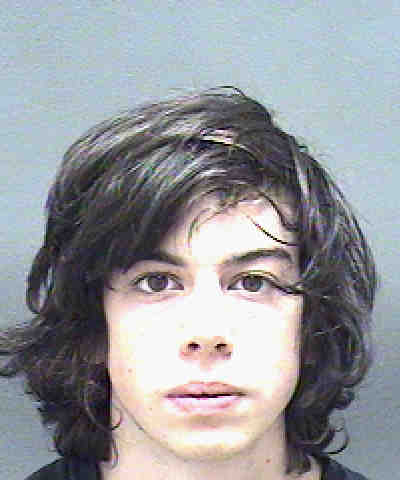}&
\includegraphics[height=1.0cm]{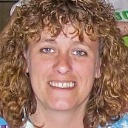}&
\includegraphics[height=1.0cm]{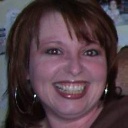}&
\includegraphics[height=1.0cm]{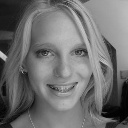}&
\includegraphics[height=1.0cm]{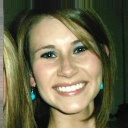}&
\includegraphics[height=1.0cm]{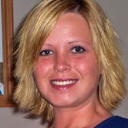}&
\includegraphics[height=1.0cm]{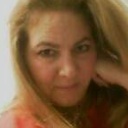}\\
{\bf Prediction} &{\color{ForestGreen} 57.0}&{\color{ForestGreen} 41.0} &{\color{ForestGreen}29.0}&{\color{red}31.1}&{\color{red}35.4}&{\color{red}39.4}&{\color{ForestGreen}Male}&{\color{ForestGreen}Female}&{\color{red}Male}&{\color{red}Female}&{\color{ForestGreen}-0.88}&{\color{ForestGreen}-0.58}&{\color{ForestGreen}0.80}&{\color{red}0.21 }&{\color{red}-0.56}&{\color{red}0.14}  \\ 
{\bf Ground truth} &57&41&29&45&55&23&Male&Female&Female&Male&-0.96&-0.59&0.79&2.68&2.60&-2.89 

\end{tabular}
\end{center}
\vspace*{-0.2cm}
\caption{Examples of accurately and wrongly predicted age, gender, and facial beauty for the MORPH 2 and Gray datasets.}
\vspace*{-0.4cm}
\label{fig:agegenderbeautyexamples}
\end{figure*}

As shown in Table~\ref{tab:hotness} our proposed features achieve state-of-the-art performance on predicting facial beauty as averaged over multiple raters. We improve by more than 10\% over the best score reported by~\cite{gray2010predicting} for the raw images. A couple of per image results are depicted in Fig.~\ref{fig:agegenderbeautyexamples}.

%-----------------------------------------------------------------------------------------

\vspace*{-0.2cm}
\section{Predicting preferences}
\label{sec:method}
\vspace*{-0.1cm}

%In Section~\ref{sec:demographics} we discuss our features and proposed methods for estimating age, gender, and average facial beauty ratings. 
Our goal is to make personalized predictions, such as how a specific male subject $m\in M$ rates a female subject $f\in F$.
The rating $R_{mf}$ is 1 if `$m$ likes $f$', -1 if `$m$ dislikes $f$', and 0 if unknown.
$f$ is also called the query user, as at test time we want to predict the individual ratings of all men for that woman. Due to space limitations, we derive the formulation for this case. Yet it is also valid when swapping sexes, \ie when women are rating men.

In the following section we phrase the problem as a collaborative filtering problem, assuming that we know past ratings for both men and women.
In Section~\ref{sec:visreg} we extend the formulation to also consider the visuals of the subjects being rated.
In Section~\ref{sec:visquery} we present a solution to predict the ratings solely based on the visual information of the subjects, without knowing how they were rated in the past.

\subsection{Model-based collaborative filtering (MF)}
\label{sec:mf}
We phrase the problem of a user $m$ rating the image of user $f$ as a model-based collaborative filtering problem. The model learned from known ratings is then used to predict unknown ratings. In its most general form, we have
\begin{equation}
\vspace{-0.1cm}
g(P_m,Q_f)\!\Rightarrow\! R_{mf},~m\!=\!1,2,...,M,~f\!=\!1,2,...,F,
\vspace{-0.1cm}
\end{equation}
where the function $g$ maps the model parameters to the known ratings.
$P_m$ denotes a set of model parameters describing the preferences of user $m$.
Similarly, $Q_f$ describes the appearance of user $f$, \ie a low-dimensional representation of how the appearance of a user is perceived.
We now estimate the model parameters given the ratings we know.

In recent years, Matrix Factorization (MF) techniques have gained popularity, especially through the Netflix challenge, where it achieved state-of-the-art performance~\cite{koren2009matrix}.
The basic assumption underlying MF models is that we can learn low-rank representations, so-called latent factors, to predict missing ratings between user $m$ and image $f$. One can approximate the ratings as
\begin{equation}
\label{eq:mfbasics}
\vspace{-0.1cm}
R\approx P^TQ=\hat{R}.
\vspace{-0.1cm}
\end{equation}
In the most common formulation of MF~\cite{shi2014collaborative} we can then frame the minimization as 
\begin{equation}
\vspace{-0.1cm}
\begin{array}{ll}
\!\!\!\!\!\!\!\!\!\!\!P^{*},Q^{*}\!=&\!\!\!\!\!\underset{P,Q}{\operatorname{argmin}}\frac{1}{2} \sum_{m=1}^M \sum_{f=1}^FI_{mf}(R_{mf}\!-\!\!P_m^TQ_f)^2\\
&\!\!\!\!\!+\!\frac{\alpha}{2}(\norm{P}^2\!+\!\norm{Q}^2)
\end{array}
\label{eq:mf}
\vspace{-0.1cm}
\end{equation}
where $P$ and $Q$ are the latent factors and $P^*$ and $Q^*$ their optimal values.
$I_{mf}$ is an indicator function that equals $1$ if there exists a rating $R_{mf}$.
The last term regularizes the problem to avoid overfitting.

\subsection{Visual regularization (MF+VisReg)}
\label{sec:visreg}
Knowing that the users in the app rate the subjects of the opposite sex solely based on the image, we make the assumption that people with similar visual features have similar latent appearance factors $Q$. Thus we can extend the formulation by adding the visual features $V$ of the query images to further regularize the optimization
\begin{equation}
\label{eq:mfvisuals}
\vspace{-0.1cm}
\begin{array}{ll}
L(P,Q)\!=\!&\!\frac{1}{2} \sum_{m=1}^M \sum_{f=1}^FI_{mf}(R_{mf}-P_m^TQ_f)^2\\
&\!+\!\frac{\alpha_1}{2}(\norm{P}^2+\norm{Q}^2)\\
&\!+\!\frac{\alpha_2}{2}\sum_{f=1}^F\sum_{g=1}^F(S_{fg}-Q_f^TQ_g)^2.
\end{array}
\vspace{-0.1cm}
\end{equation} 
The visual similarity is defined as 
\begin{equation}
\vspace{-0.1cm}
\label{eq:vissim}
%S_{fg}=\left(\frac{V_{f}^TV_{g}}{\norm{V_{f}}\norm{V_{g}}}\right)^\gamma,
S_{fg}=\frac{V_{f}^TV_{g}}{\norm{V_{f}}\norm{V_{g}}}.
\vspace{-0.1cm}
\end{equation}
Visually this proves to be a good metric for visual similarity. 
The optimal latent factors are calculated by gradient descent, where the derivatives are
\begin{equation}
\vspace{-0.1cm}
\label{eq:gradients}
\begin{array}{ll}
\frac{\partial L}{\partial P_m}=&\sum_{f=1}^FI_{mf}(P_m^TQ_f-R_{mf})Q_f+\lambda P_m\\
\frac{\partial L}{\partial Q_f}=&\sum_{m=1}^MI_{mf}(P_m^TQ_f-R_{mf})P_m\\
&+2\alpha_2 \sum_{g=1}^F(Q_f^TQ_g-S_{fg})Q_g+\lambda Q_f.
\end{array}
\vspace{-0.1cm}
\end{equation}

\subsection{Visual query}
\label{sec:visquery}
We now want to predict how user $m$ rates user $f$ without knowing any past ratings of $f$ but knowing her visual feature $V_f$ (see Fig.~\ref{fig:pipeline}). This implies that we do not know the latent factor $Q_f$ for $f$.
The goal is to get an estimate $\hat{Q}_f$ of $Q_f$ based solely on the visual feature $V_f$.
Then we would be able to regress the rating as 
\begin{equation}
\vspace{-0.1cm}
\label{eq:ratingestimate}
\hat{R}_{mf}=P_m^T\hat{Q}_f.
\vspace{-0.1cm}
\end{equation}
Learning a global regression led to poor results as attractiveness is highly subjective. Instead our approach is inspired by the recently introduced anchored neighborhood regression (ANR) method for image super-resolution~\cite{Timofte-ACCV-2014}, where the problem is formulated as a piece-wise local linear regression of low to high resolution image patches and with offline trained regressors. In contrast to ANR, each sample is an anchor and the neighborhood is spanned over all other training samples and weighted according to its similarity to the anchor. This way we are obtaining more robust local regressors that can cope with the scarcity of the data.

As for regularizing MF, we assume that the visual features $V$ and the latent factor $Q$ locally have a similar geometry. Further, we assume that we can locally linearly reconstruct each visual feature or latent factor by its neighbors. 
Under these assumptions we can reconstruct features and latent factors using the same weights for the neighbors. In the visual space we now aim to find these weights $\beta$ by phrasing the problem as a ridge regression
\begin{equation}
\vspace{-0.1cm}
\underset{\beta_g}{\operatorname{min}}\norm{V_g\!-\!N_{V_g}\beta_g}^2\!+\!\lambda\left(\kappa\norm{\Gamma_g \beta_g}^2\!+\!(1\!-\!\kappa)\norm{\beta_g}^2\right),
\vspace{-0.1cm}
\end{equation}
where $N_{V_g}$ is a matrix of the neighboring visual features of $V_g$ stacked column-wise and $\kappa$ is a scalar parameter. The optimization is regularized by the similarity to its neighbors according to eq.~\ref{eq:vissim}, in the sense that greater similarity yields greater influence on $\beta$:
\begin{equation}
\vspace{-0.1cm}
\Gamma_g=\operatorname{diag}(1-S_{g1},1-S_{g2},...,1-S_{gF}).
\vspace{-0.1cm}
\end{equation}
The closed-form solution of the problem can be written as 
\begin{equation}
\beta_k\!=\!\left[N_{V_g}^TN_{V_g}\!+\!\lambda\left[\kappa\Gamma_g^T\Gamma_g\!+\!(1\!-\!\kappa)I\right]\right]^{-1}N_{V_g}^TV_g.
\end{equation}
As we assume that the latent space behaves similarly locally, we can regress the latent factor $Q_g$ as a linear combination of its neighbors using the same $\beta_g$. Note that $N_{Q_g}$ corresponds to the latent factors of $N_{V_g}$, \ie the neighbors in the visual space.
Plugging in our solution for $\beta_g$ we get
\begin{equation}
\vspace{-0.1cm}
\begin{array}{ll}
\!\!\!\!\!\!\!\!Q_g\!&\!\!\!\!\!=\!N_{Q_g}\beta_g\\
&\!\!\!\!\!=\!N_{Q_g}\!\!\left[N_{V_g}^TN_{V_g}\!\!+\!\lambda\left[\kappa\Gamma_g^T\Gamma_g\!+\!(1\!-\!\kappa)I\right]\right]^{-1}\!\!\!\!\!\!\!N_{V_g}^TV_g\\
&\!\!\!\!\!=\!M_{g}V_g.
\end{array}
\vspace{-0.1cm}
\end{equation}
Thus we have found a projection $M_g$ from a visual feature $V_g$ to its latent factor $Q_g$.
At test time for a given visual feature $V_f$, we now aim to find the most similar visual feature in the training space, $\hat{g}=\arg\max_{g}~S_{fg}$.
Then we use the projection matrix of $\hat{g}$ to obtain $\hat{Q}_f$ to finally estimate the rating of user $m$ for the image of user $f$ as
\begin{equation}
\vspace{-0.1cm}
\label{eq:ratingestimate2}
\hat{R}_{mf}=P_m^T\hat{Q}_f,\quad \hat{Q}_f=M_{\hat{g}}V_f.
\vspace{-0.1cm}
\end{equation}

\vspace*{-0.2cm}
\section{Experiments}
\label{sec:experiments}
\vspace*{-0.1cm}
\noindent In this section we present qualitative and quantitative results of our proposed framework on the Hot-or-Not and the MovieLens dataset.
\vspace*{-0.1cm}
\subsection{Hot-or-Not}
\vspace*{-0.1cm}
\subsubsection{The dataset}
\vspace*{-0.1cm}
\label{sec:dataset}
Our dataset was kindly provided by Blinq\footnote{www.blinq.ch}, a popular hot-or-not dating application. We will make the anonymized ratings and visual features of the last layer available under \\ \url{http://www.vision.ee.ethz.ch/~rrothe/}.
The app shows the user people of the sex of interest, one after the other. The user can then like or dislike them. If both like each other's profile photo they are {\it matched} and can chat to each other. People can select up to 5 photos from Facebook for their profile.

\noindent{\bf{Dataset statistics.}}
Before performing any experiments we removed underage people, anyone over 37 and bi- and homosexual users as these comprise only a small minority of the dataset. All users who received less than 10 ratings were also removed. As the majority of the people decide on the first photo, we ignore the other photos.
The resulting dataset has 4,650 female users and 8,560 male users.
The median age is 25.
In total there are 17.33 million ratings, 11.27m by men and 6.05m by women.
Interestingly, 44.81\% of the male ratings are positive, while only 8.58\% of the female ratings are positive. 
Due to this strong bias of ratings by women, we only predict the ratings of men.
There are 332,730 matches.
%The dataset is relatively sparse, \ie only 28.29\% of the ratings of men rating women are available, and 15.14\% vice versa.

\begin{figure}[tbph!]
\vspace*{-0.2cm}
\begin{center}
\setlength{\tabcolsep}{0pt}
\begin{tabular}{cc}
{\setlength{\tikzpicturewidth}{2cm}\setlength{\tikzpictureheight}{2cm}\input{Figures/age_stats_gender_1.tikz}}&
{\setlength{\tikzpicturewidth}{2cm}\setlength{\tikzpictureheight}{2cm}\input{Figures/age_stats_gender_0.tikz}}\\
\end{tabular}
\end{center}
\vspace*{-0.5cm}
\caption{Preferences by age for women and men. %\textcolor{red}{Red} indicates that the age group likes the respective other group.
}
\vspace*{-0.25cm}
\label{fig:prefbyage}
\end{figure}
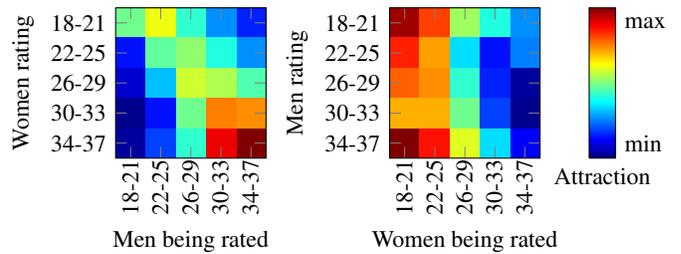

\noindent{\bf{Preferences bias.}}
To investigate the natural bias caused by age, we divide the men and women from our dataset according to their age and gender. For each age group of men we counted the percent of hot vs. not on the ratings towards the age groups of women and vice versa. Fig.~\ref{fig:prefbyage} describes the preferences by age as found in our dataset. Women generally prefer slightly older men and give better ratings the older they get. In comparison, men on this app prefer women under 25. 

\begin{figure}[tbph!]
\vspace*{-0.3cm}
\begin{center}
{\setlength{\tikzpicturewidth}{6cm}\setlength{\tikzpictureheight}{2.5cm}\input{Figures/hotness_paradox.tikz}}
\end{center}
\vspace*{-0.3cm}
\caption{{\bf Hotness paradox.} The people visually similar to you are on average hotter than you.
The situation changes when we compute the similarity based on learned latent Q representations.}
\vspace*{-0.3cm}
\label{fig:hotness_paradox}
\end{figure}
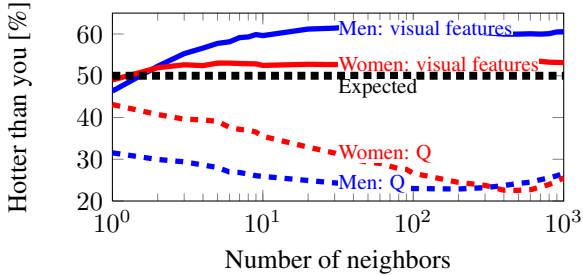

\noindent{\bf{Hotness paradox.}} We notice an interesting phenomenon.
Most people have a lower rating than their visually similar neighbors, on average, and this holds for both men and women. In Fig.~\ref{fig:hotness_paradox} we report the percentage of cases where the average hotness of the neighborhood of a subject is greater than that of the subject, and this for a neighborhood size from 1 to $10^3$. We plot also the results when we use our latent Q representations for retrieving neighbors.
Surprisingly, this time the situation is reversed, the subjects tend to be hotter than their Q-similar neighborhood. Regardless the choice of similarity we have a strong deviation from the expected value of $50\%$.
We call this phenomenon the `Hotness paradox'. It relates to the so-called `Friendship paradox'~\cite{Feld-AJS-1991} in social networks, where most people have fewer friends than their friends have.

\noindent{\bf{Visual features.}} As the space covered by the person varies greatly between images, we run a top face detector~\cite{mathias2014face} on each image.
Then we crop the image to the best scoring face detection and include its surrounding (100\% of the width to each side and 50\% of the height above and 300\% below), to capture the upper-body and some of the background. If the face is too small or the detection score too low, we take the entire image. Then we extract CNN features.
\vspace*{-0.25cm}
\subsubsection{Experimental setup}
\vspace*{-0.1cm}
For all experiments, $50\%$ of either gender are used for training and the rest for testing.
For each user in the testing set, $50\%$ of the received ratings are used for testing.
We compare different methods. {\it Baseline} predicts the majority rating in the training set.
Matrix factorization is applied without and with visual regularization, {\it MF} ($\alpha_1=0.1$) and {\it MF+VisReg} ($\alpha_2=0.1$), resp.
The dimensionality of the latent vector of $P$ and $Q$ is fixed to 20.
The other parameters were set through cross-validation on a subset of the training data.
We predict the ratings a subject receives based upon different knowledge:
For {\it Visual} we solely rely on the image of the subject which means that we do not know any ratings the subject has received so far.
For {\it 10 Ratings, 100 Ratings, Full History}, we instead base the prediction upon a fixed set of known ratings for each query user.
We report the average accuracy, \ie the percentage of correctly predicted ratings of the testing set, and the Pearson's correlation.
\vspace*{-0.25cm}
\subsubsection{Results}
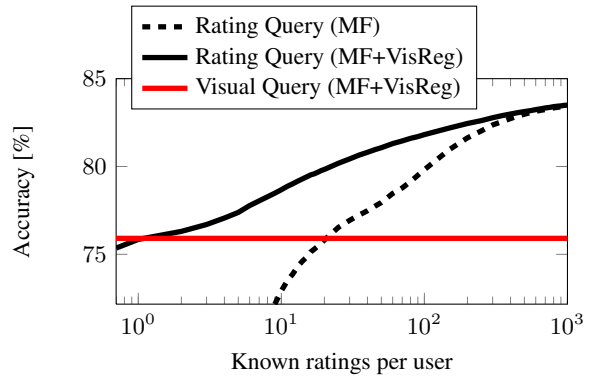
\begin{figure}[th!]
\vspace*{-0.2cm}
\begin{center}
{\setlength{\tikzpicturewidth}{6cm}\setlength{\tikzpictureheight}{3cm}\input{Figures/nr_ratings_vs_acc.tikz}}
\end{center}
\vspace*{-0.4cm}
\caption{Number of known ratings for a female query user vs. accuracy of predicted male's ratings.}
\vspace*{-0.4cm}
\label{fig:nrratingsvsacc}
\end{figure}

\vspace*{-0.1cm}
\noindent{\bf{Performance.}} Fig.~\ref{fig:nrratingsvsacc} shows how the average accuracy varies with the number of known past ratings for the query user.
We report the average performance across all men's ratings. Knowing just the image of the person, we can predict $75.92\%$ of the ratings correctly. Adding past received ratings of the user improves performance to up to $83.64\%$. 
Matrix factorization significantly improves as more ratings are known. 
If only few ratings are available, regularizing the matrix factorization with the visuals boosts performance significantly, \ie from $72.92\%$ to $78.68\%$ for 10 known ratings.
Table~\ref{tab:Hot-or-Not} summarizes the results for various settings.

\begin{table}[ht!]
\scriptsize
\vspace*{-0.2cm}
\begin{center}
\begin{tabular}{c|ccc}
            & Query         & Accuracy      & Correlation\\
\hline
Baseline    & N/A           & 54.92\%       & N/A\\ %0.902
\hdashline
%Mean         & 74.96\%        & 0.501   & 0.506 \\
%PCA          & 75.56\%        & 0.489   & 0.514\\
MF          & \multirow{2}{*}{Visual}        & 75.90\% & 0.520\\ %{\bf 0.482} 
MF+VisReg   &         & {\bf 75.92\%} & {\bf0.522}\\ 
\hline
MF          &   \multirow{2}{*}{10 Ratings}   &    72.92\%         & 0.456\\
MF+VisReg   &     &    78.68\%       & 0.576  \\
\hdashline
MF          & \multirow{2}{*}{100 Ratings}     & 79.82\%      & 0.593\\
MF+VisReg   &     &    81.82\%   &  0.635\\
\hdashline
 MF         & \multirow{2}{*}{Full History}   &  83.62\%       & 0.671\\ %0.328 
MF+VisReg   &   &    {\bf 83.64\%}    &  {\bf 0.671}\\
%\hline
\end{tabular}
\end{center}
\vspace*{-0.2cm}
\caption{Preference prediction results on Hot-or-Not dataset for female queries.}
\vspace*{-0.2cm}
\label{tab:Hot-or-Not}
\end{table}

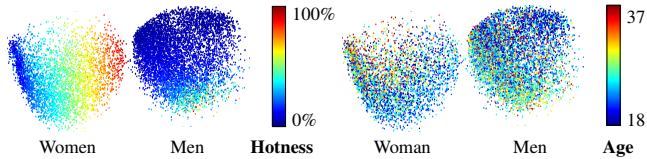
\begin{figure}[th!]
\scriptsize
\setlength{\tabcolsep}{0pt}
\vspace*{-0.2cm}
\begin{center}
\begin{tabular}{cccc}
{\setlength{\tikzpicturewidth}{1.6cm}\setlength{\tikzpictureheight}{1.6cm}\input{Figures/hotness_q_gender_0.tikz}}&
{\setlength{\tikzpicturewidth}{1.6cm}\setlength{\tikzpictureheight}{1.6cm}\input{Figures/hotness_q_gender_1.tikz}}&
{\setlength{\tikzpicturewidth}{1.6cm}\setlength{\tikzpictureheight}{1.6cm}\input{Figures/age_q_gender_0.tikz}}&
{\setlength{\tikzpicturewidth}{1.6cm}\setlength{\tikzpictureheight}{1.6cm}\input{Figures/age_q_gender_1.tikz}}\\
 Women &~~~Men~~~~~~~~~~{\bf Hotness}& Woman &~~~~~~~Men~~~~~~~~~~~~{\bf Age}\\
\end{tabular}
\end{center}
\vspace*{-0.2cm}
\caption{Visualization of latent space Q for women and men.}
\vspace*{-0.2cm}
\label{fig:visualizeq}
\end{figure}

\noindent{\bf{Latent space $Q$ vs. preferences.}} In Fig.~\ref{fig:visualizeq} we show the learned latent space $Q$ from the matrix factorization by PCA projecting it to two dimensions and adding the hotness and age properties for both genders with visual regularization. 
The learned latent factor $Q$ captures appearance and for women there is a clear separation in terms of attractiveness and age, whereas for men the separation is less obvious.

\begin{figure*}[]
\begin{center}
\setlength{\tabcolsep}{1pt}
\scriptsize
\begin{tabular}{lcccc|cccc|cccc}
 &  No filter &  Earlybird &  X-Pro II &  Valencia &  No filter &  Earlybird &  X-Pro II &  Valencia &  No filter &  Earlybird &  X-Pro II &  Valencia\\
&
\includegraphics[height=1.3cm]{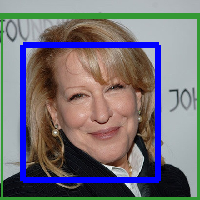}&
\includegraphics[height=1.3cm]{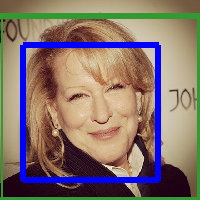}&
\includegraphics[height=1.3cm]{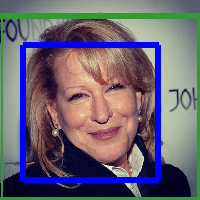}&
\includegraphics[height=1.3cm]{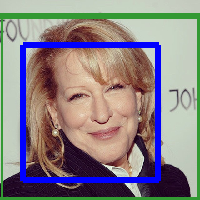}&
\includegraphics[height=1.3cm]{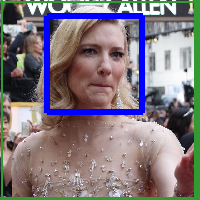}&
\includegraphics[height=1.3cm]{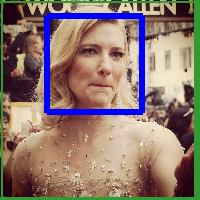}&
\includegraphics[height=1.3cm]{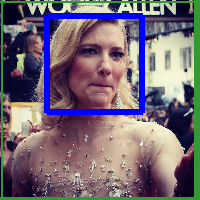}&
\includegraphics[height=1.3cm]{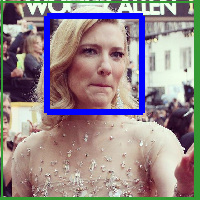}&
\includegraphics[height=1.3cm]{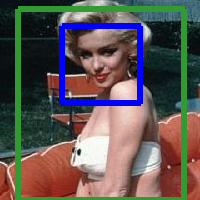}&
\includegraphics[height=1.3cm]{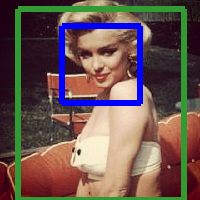}&
\includegraphics[height=1.3cm]{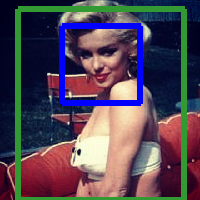}&
\includegraphics[height=1.3cm]{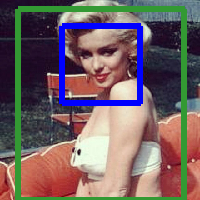}\\
& 26\%  & {\bf42\%} & 25\%& 40\%  & 19\%  & 40\% & {\bf 41\%}& 31\%  & 32\%  & 41\% & 46\%& {\bf59\%}  \\
\end{tabular}
\end{center}
\vspace{-0.3cm}
\caption{Improving the hotness rating by Instagram filters.}
\label{fig:instagram}
\vspace{-0.3cm}
\end{figure*}

%\begin{figure}[th!]
%\vspace*{-0.2cm}
%\begin{center}
%\includegraphics[width=6.5cm, height=5cm]{Figures/images_q_gender_0.png}
%\end{center}
%\vspace*{-0.2cm}
%\caption{Female images mapped into latent space Q.}
%\vspace*{-0.2cm}
%\label{fig:qimages}
%\end{figure}

\begin{figure}[th!]
\vspace*{-0.2cm}
\begin{center}
\includegraphics[width=1\linewidth]{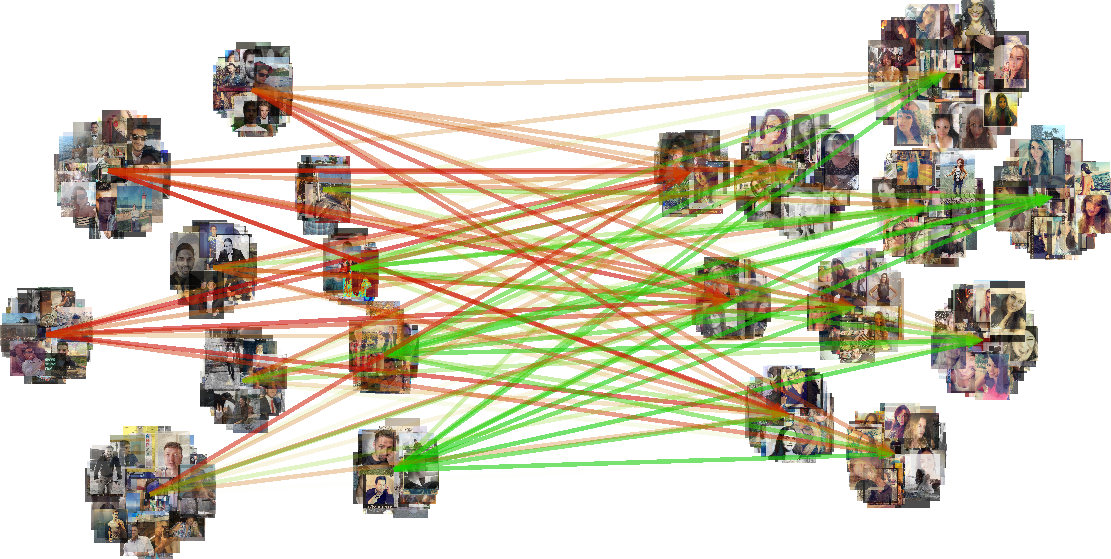}
\end{center}
\vspace*{-0.2cm}
\caption{Preferences between clusters of users. The color of the arrow indicates how much the men's cluster likes (\textcolor{ForestGreen}{green}) or dislikes (\textcolor{red}{red}) the women's cluster on average. }
\vspace*{-0.4cm}
\label{fig:qcluster}
\end{figure}

According to the preferences $P$ and the learned latent $Q$ one can have a more in-depth view on the appearance of women and men. 
In Fig.~\ref{fig:qcluster} both men and women are clustered according to their 2D representation of the learned latent factors $P$ (preferences of men) and $Q$ (appearances of women), respectively. For visualization purposes we used 100 user images for each cluster and 10 clusters. The men are visually more diverse in each of their clusters than the women in their clusters, because the men are clustered according to their preferences, therefore ignoring their visual appearance, while the women are clustered according to their Q factors which are strongly correlated with appearance and hotness, as shown in Fig.~\ref{fig:visualizeq}.

\begin{figure*}[th!]
\begin{center}
\setlength{\tabcolsep}{1pt}
%\small
\scriptsize
\begin{tabular}{lccc|ccc|cccc}
&  \multicolumn{3}{c|}{Helena Bonham Carter} &  \multicolumn{3}{c|}{Natalie Portman}&  \multicolumn{4}{c}{Charlize Theron}  \\
&
\includegraphics[height=1.5cm]{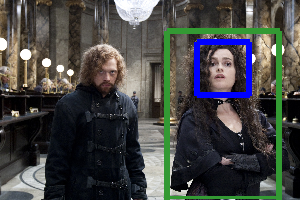}&
\includegraphics[height=1.5cm]{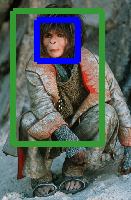}&
\includegraphics[height=1.5cm]{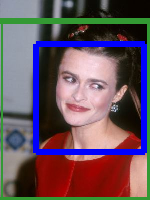}&
\includegraphics[height=1.5cm]{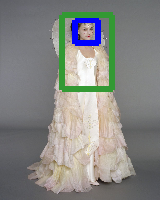}&
\includegraphics[height=1.5cm]{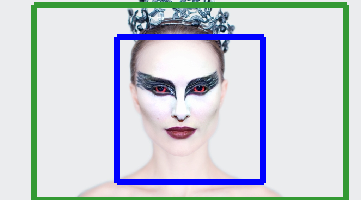}&
\includegraphics[height=1.5cm]{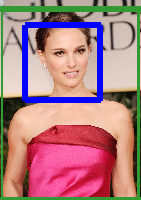}&
\includegraphics[height=1.5cm]{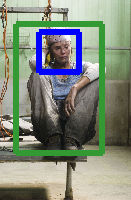}&
\includegraphics[height=1.5cm]{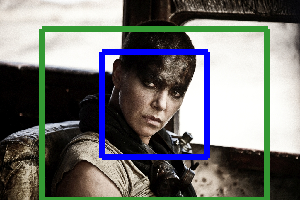}&
\includegraphics[height=1.5cm]{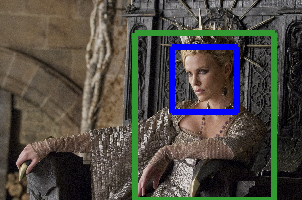}&
\includegraphics[height=1.5cm]{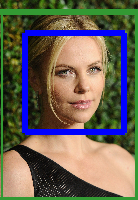}\\
& 20\%  & 35\% & 62\% & 19\% & 35\% & 59\% & 27\% & 31\%&46\% &68\%  \\
\end{tabular}
\begin{tabular}{lccc|ccc|ccc}
&  \multicolumn{3}{c|}{Cate Blanchett}&  \multicolumn{3}{c|}{Bette Midler}&  \multicolumn{3}{c}{Jim Carrey}  \\
&
\includegraphics[height=1.51cm]{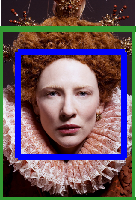}&
\includegraphics[height=1.51cm]{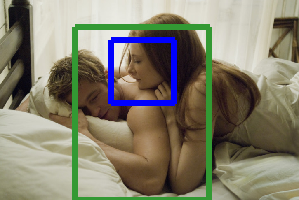}&
\includegraphics[height=1.51cm]{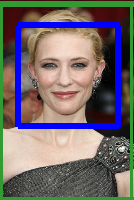}&
\includegraphics[height=1.51cm]{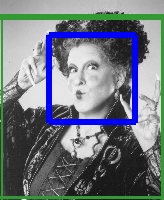}&
\includegraphics[height=1.51cm]{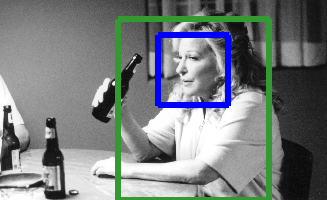}&
\includegraphics[height=1.51cm]{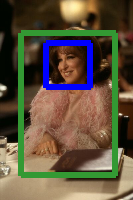}&
\includegraphics[height=1.51cm]{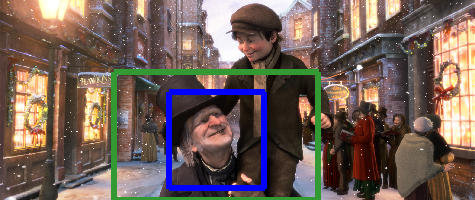}&
\includegraphics[height=1.51cm]{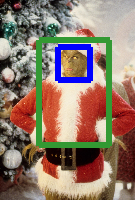}&
\includegraphics[height=1.51cm]{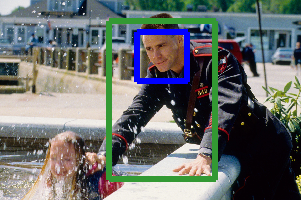}\\
& 26\%  & 42\% & 59\% & 20\% & 32\% & 47\% & 19\% & 32\%&59\%  \\
\end{tabular}
\begin{tabular}{lcc|cc|cc|cccc}
&  \multicolumn{2}{c|}{Cats}&  \multicolumn{2}{c|}{Dogs}&  \multicolumn{2}{c|}{Wonder Woman}&  \multicolumn{4}{c}{Some like it hot}  \\
&
\includegraphics[height=1.5cm]{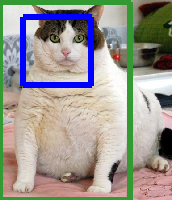}&
\includegraphics[height=1.5cm]{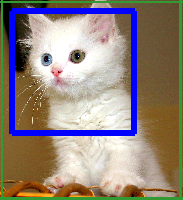}&
\includegraphics[height=1.5cm]{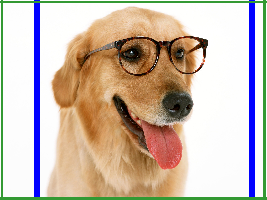}&
\includegraphics[height=1.5cm]{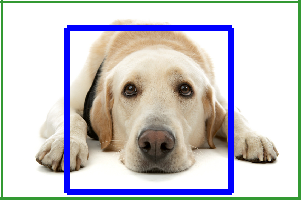}&
\includegraphics[height=1.5cm]{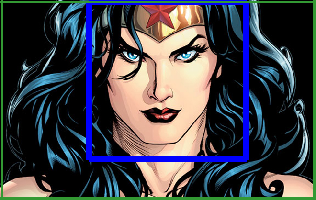}&
\includegraphics[height=1.5cm]{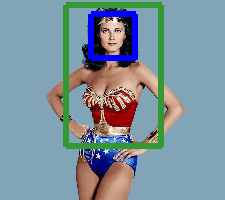}&
\includegraphics[height=1.5cm]{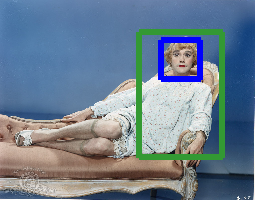}&
\includegraphics[height=1.5cm]{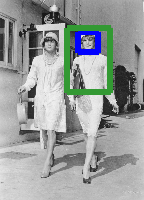}&
\includegraphics[height=1.5cm]{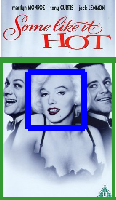}&
\includegraphics[height=1.5cm]{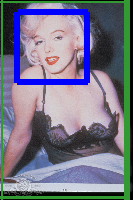}\\
& 29\%  & 33\% & 32\% & 36\% & 33\% & 47\% & 27\% & 39\%&54\% &54\%  \\
\end{tabular}
\begin{tabular}{lcc|cc|cc|cc|cc}
&   \multicolumn{2}{c|}{Melissa McCarthy}& \multicolumn{2}{c|}{Too small face}&  \multicolumn{2}{c|}{Face detector fails}& \multicolumn{2}{c|}{Wrong person}&  \multicolumn{2}{c}{Wrong prediction}\\
&
\includegraphics[height=1.41cm]{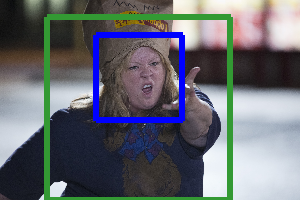}&
\includegraphics[height=1.41cm]{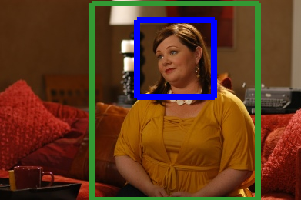}&
\includegraphics[height=1.41cm]{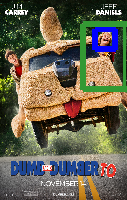}&
\includegraphics[height=1.41cm]{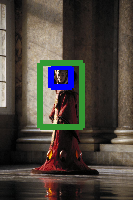}&
\includegraphics[height=1.41cm]{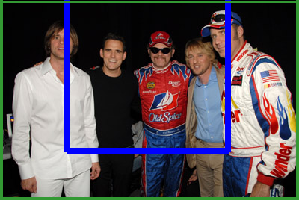}&
\includegraphics[height=1.41cm]{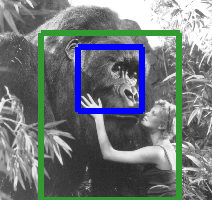}&
\includegraphics[height=1.41cm]{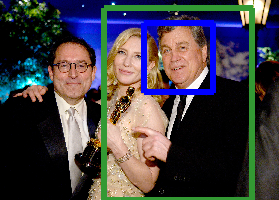}&
\includegraphics[height=1.41cm]{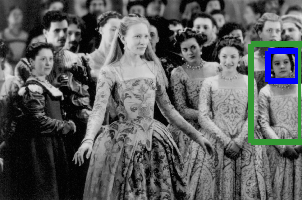}&
\includegraphics[height=1.41cm]{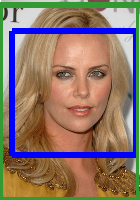}&
\includegraphics[height=1.41cm]{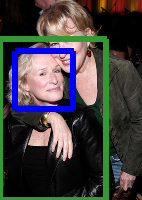}\\
& 24\%  & 45\% & 32\% & 39\% & 34\% & 39\% & 14\% & 38\%&18\% &67\%  \\
\end{tabular}
\end{center}

\vspace*{-0.2cm}
\caption{Predicted percentage of positive ratings for numerous celebrities by the user base of the Hot-or-Not dataset.}
\vspace*{-0.2cm}
\label{fig:celebrities}
\end{figure*}

\noindent{\bf{Visual queries without past ratings.}} We validate our approach on images outside our dataset, retrieved from the internet for celebrities. By applying the visual query regression to the Q space we can make good predictions for such images. For a visual assessment see Fig.~\ref{fig:celebrities}.
This figure also depicts a number of issues our pipeline faces: too small faces, detector failure, wrongly picked face, or simply a wrong prediction.
We also tested our method on cartoons and companion pets with the predictor trained on Hot-or-Not.
The results are surprising.

\noindent{\bf{Instagram filters or how to sell your image.}}
Images and their hotness prediction also indicate which changes could improve their ratings. Earlier work has aimed at the beautification of a face image by invasive techniques such as physiognomy changes~\cite{Leyvand-SIG-2008} or makeup~\cite{Guo-CVPR-2009,Liu-MCCA-2014}. Yet, we found that non-invasive techniques (not altering facial geometry and thus `fair') can lead to surprising improvements. We have evaluated the most popular Instagram filters\footnote{brandongaille.com/10-most-popular-instagram-photo-filters} for our task. We observed that the filters lead to an increase in predicted hotness. In Fig.~\ref{fig:instagram} we show a couple of results in comparison to the original image.
Note that with our predictor and such Instagram filters a user can easily pick its best profile photo.

\vspace*{-0.25cm}
\subsubsection{howhot.io}
\vspace*{-0.1cm}
\begin{wrapfigure}{r}{0.3\linewidth}
\vspace{-1cm}
  \begin{center}
    \includegraphics[width=1\linewidth]{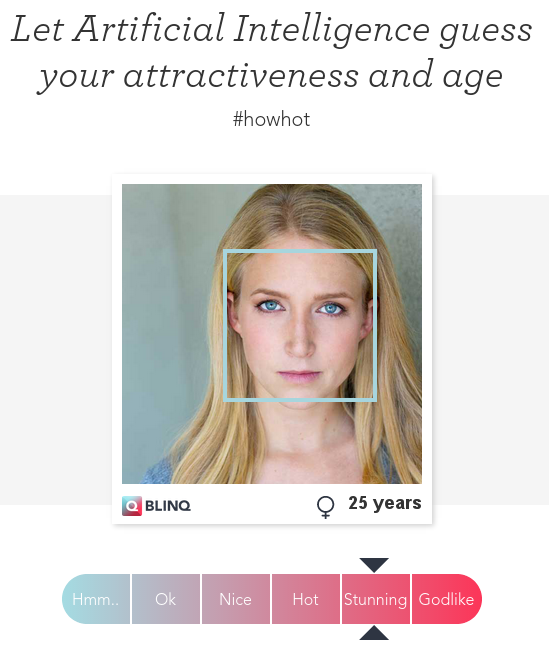}
  \end{center}
  \vspace{-0.2cm}
\caption{howhot.io}
\vspace{-0.4cm}
\label{fig:howhot}
\end{wrapfigure}

We demonstrate our algorithms on \url{howhot.io}, a website where people can upload a photo of their face and our algorithm will then estimate the age, gender and facial attractiveness of the person (c.f. Fig.~\ref{fig:howhot}). The CNN was trained on the Hot-or-Not dataset for predicting attractiveness and on the IMDB-WIKI dataset~\cite{Rothe-ICCVW-2015} for age and gender prediction.
The website went viral around the Internet with more than 50 million pictures evaluated in the first month.

\vspace*{-0.1cm}
\subsection{MovieLens}
\subsubsection{The dataset}
\vspace*{-0.1cm}
We also perform experiments on the MovieLens 10M\footnote{grouplens.org/datasets/movielens} dataset.
It contains 10,000,054 ratings from 69,878 users for 10,681 movies.
Ratings are made on a 5-star scale, with half-star increments.
On average, each movie has 3,384 ratings and each user rates 517 movies.
Note that even though there are more than 10 million ratings, the rating matrix is sparse with only 1.34\% of all ratings known.
We augment each movie with the poster image from IMDB and extract the same deep CNN features as for the Hot-or-Not dataset. 
We will make the poster images publicly available under\\ \url{http://www.vision.ee.ethz.ch/~rrothe/}.

\vspace*{-0.25cm}
\subsubsection{Experimental Setup}
\vspace*{-0.1cm}
The experimental setup in term of training and testing split is identical to the Hot-or-Not dataset.
As the movie posters are much less informative regarding the ratings in comparison to the Hot-or-Not images, the visual regularization is reduced to $\alpha_2=0.001$.
For a given movie we want to infer the ratings of all users.
Again, we evaluate the case where just the poster is known and also cases where a varying number of ratings is known.
As a baseline we show how a prediction made at random would perform,
assuming that there is no bias in the ratings of the test set.

\begin{figure*}[tbph!]
\vspace*{-0.1cm}
\begin{center}
\setlength{\tabcolsep}{1pt}
\scriptsize
\begin{tabular}{lcccccc@{\hskip 0.5cm}ccc@{\hskip 0.5cm}ccc}
&\multicolumn{6}{c}{Correctly predicted}  &\multicolumn{3}{c}{Overrated poster}   & \multicolumn{3}{c}{Underrated poster} \\
&
\includegraphics[height=1.5cm]{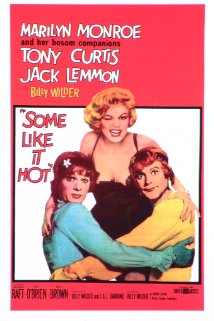}&
\includegraphics[height=1.5cm]{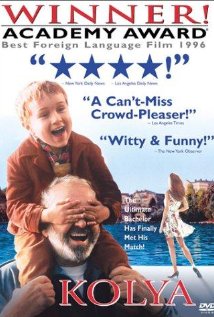}&
\includegraphics[height=1.5cm]{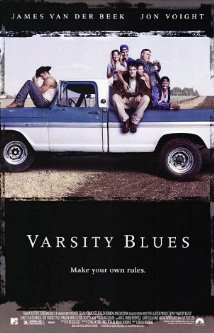}&
\includegraphics[height=1.5cm]{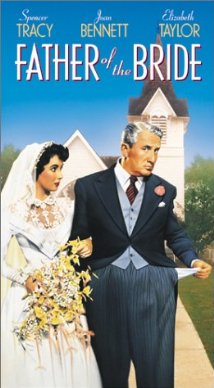}&
\includegraphics[height=1.5cm]{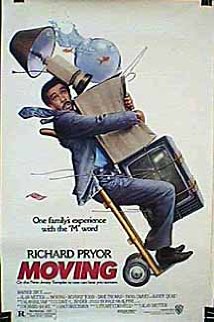}&
\includegraphics[height=1.5cm]{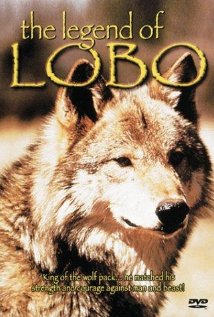}&
\includegraphics[height=1.5cm]{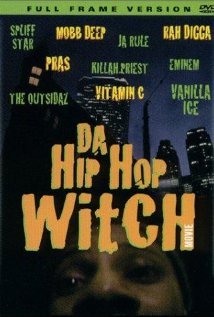}&
\includegraphics[height=1.5cm]{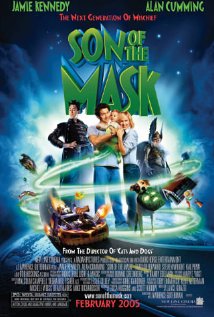}&
\includegraphics[height=1.5cm]{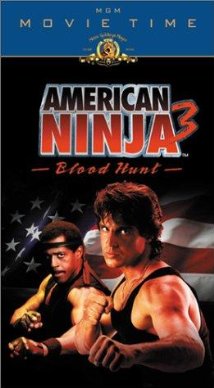}&
\includegraphics[height=1.5cm]{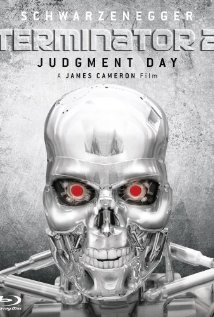}&
\includegraphics[height=1.5cm]{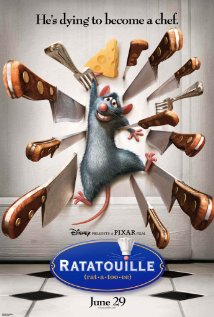}&
\includegraphics[height=1.5cm]{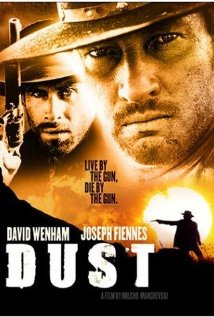}\\
%& \tiny Kolja & \tiny Varsity Blues & \tiny Father of the Bride & \tiny Moving & \tiny Da Hip Hop Witch & \tiny Son of the Mask & \tiny Terminator 2 & \tiny Movie H \\
{\bf Average prediction} & {\color{ForestGreen}3.9} & {\color{ForestGreen}4.0} & {\color{ForestGreen}3.1} & {\color{ForestGreen}3.5} & {\color{ForestGreen}2.7} & {\color{ForestGreen}2.4} & {\color{blue}3.8} & {\color{blue}3.5} & {\color{blue}3.7} & {\color{red}2.4} & {\color{red}2.9}& {\color{red}2.7} \\
{\bf Average rating} & 4.1 & 4.0 & 3.1 & 3.5 & 2.7 & 2.5 & 0.6 & 1.4 & 1.9 & 3.9 & 3.9 &4.5\\
\end{tabular}
\end{center}
\vspace*{-0.2cm}
\caption{Examples of predicted ratings for various movie posters solely based on the visual information of the poster.}
\vspace*{-0.2cm}
\label{fig:movieposterexamples}
\end{figure*}

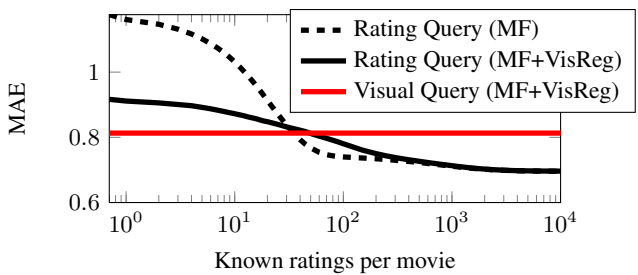
\begin{figure}[]
\vspace*{-0.2cm}
\begin{center}
{\setlength{\tikzpicturewidth}{6cm}\setlength{\tikzpictureheight}{2.5cm}\input{Figures/movielens_nr_ratings_vs_mae.tikz}}
\end{center}
\vspace*{-0.4cm}
\caption{Number of known ratings for a movie vs. MAE of the predicted ratings.}
\vspace*{-0.4cm}
\label{fig:movielensnrratingsvsmae}
\end{figure}

\begin{table}[]
\vspace*{-0.2cm}
\scriptsize
\begin{center}
\begin{tabular}{c|cccc}
            & Query                             & MAE      & Correlation\\
\hline
Baseline      & N/A                              & 1.507    & N/A        \\
%Baseline    & N/A                              & 0.805    & N/A        \\ 
\hdashline
%MF+Mean     &\multirow{4}{*}{Visual}            & 0.764    & 0.246      \\
%MF+PCA      &                                  & 0.758    & 0.263      \\
MF          & \multirow{2}{*}{Visual}      &  0.824    & 0.286      \\ 
MF+VisReg   &                              & {\bf 0.813} & {\bf 0.292}        \\
\hline
MF          &\multirow{2}{*}{10 Ratings}   & 1.031    &    0.280        \\
MF+VisReg   &                              & 0.872    &    0.270        \\
\hdashline
MF          &\multirow{2}{*}{100 Ratings}  & 0.740    &  0.467     \\
MF+VisReg   &                              & 0.780    &  0.461    \\
\hdashline
 MF         &\multirow{2}{*}{Full History} & 0.696    & 0.530      \\  
MF+VisReg   &                              & {\bf 0.696}    & {\bf 0.536}  \\
\hline
\end{tabular}
\end{center}
\vspace*{-0.2cm}
\caption{Rating prediction results on augmented MovieLens.}
\vspace*{-0.2cm}
\label{tab:movielens}
\end{table}

\vspace*{-0.25cm}
\subsubsection{Results}
\vspace*{-0.2cm}
Table~\ref{tab:movielens} summarizes the performance. 
Fig.~\ref{fig:movielensnrratingsvsmae} shows how the number of known ratings impacts the MAE. 
Visual regularization of MF improves performance, especially when few ratings are known, \ie for 10 known ratings the MAE can be reduced by $15\%$ from 1.031 to 0.872. 
When just the movie poster is known, the MAE is 0.813, which is on par with knowing 30 ratings.
Fig.~\ref{fig:movieposterexamples} shows some movie posters. 
We also show overrated and underrated posters, \ie posters where our algorithm - based on the poster - predicts a much better or worse score than the actual movie rating.

\vspace*{-0.2cm}
\section{Conclusion}
\label{sec:conclusions}
\vspace*{-0.2cm}
We proposed a collaborative filtering method for rating/preference prediction based not only on the rating history but also on the visual information. Moreover, we can accurately handle queries with short or lacking rating history. We evaluated our system on a very large dating dataset and on the MovieLens dataset augmented with poster images. To the best of our knowledge, we are the first to report on such a large dating dataset, and to show that adding weak visual information improves the rating prediction of collaborative filtering methods on MovieLens. We achieved state-of-the-art results on facial beauty, age and gender prediction and give some sociologically interesting insights.

\vspace{0.1cm}
\noindent{\textbf{Acknowledgements.}}
We thank Blinq for providing the Hot-or-Not data and NVIDIA for donating a Tesla K40 used in our research.
\vspace{-0.4cm}

%-------------------------------------------------------------------------
%\clearpage
{
%\small
\footnotesize
%\scriptsize 	
\bibliographystyle{ieee}
\bibliography{blinq_cvpr}
}

\end{document}

%% file: Figures/age_stats_gender_1.tikz
% This file was created by matlab2tikz v0.4.6 running on MATLAB 8.3.
% Copyright (c) 2008--2014, Nico Schlömer <nico.schloemer@gmail.com>
% All rights reserved.
% Minimal pgfplots version: 1.3
% 
\begin{tikzpicture}

\begin{axis}[%
width=\tikzpicturewidth,
height=\tikzpictureheight,
axis on top,
font=\small,
clip=false,
scale only axis,
xmin=0.5,
xmax=5.5,
xtick={1,2,3,4,5},
xticklabels={\empty},
xlabel style={align=center},
xlabel={\\[1ex]Men being rated},
y dir=reverse,
ymin=0.5,
ymax=5.5,
ytick={1,2,3,4,5},
yticklabels={{18-21},{22-25},{26-29},{30-33},{34-37}},
ylabel={Women rating}
]
\addplot [forget plot] graphics [xmin=0.5,xmax=5.5,ymin=0.5,ymax=5.5] {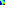};
\node[left, inner sep=0mm, rotate=90, text=black]
at (axis cs:0.98963133640553,5.529296875,1.77635683940025e-15) {18-21};
\node[left, inner sep=0mm, rotate=90, text=black]
at (axis cs:1.99193548387097,5.529296875,1.77635683940025e-15) {22-25};
\node[left, inner sep=0mm, rotate=90, text=black]
at (axis cs:2.99423963133641,5.529296875,1.77635683940025e-15) {26-29};
\node[left, inner sep=0mm, rotate=90, text=black]
at (axis cs:3.98502304147465,5.529296875,1.77635683940025e-15) {30-33};
\node[left, inner sep=0mm, rotate=90, text=black]
at (axis cs:4.98732718894009,5.529296875,1.77635683940025e-15) {34-37};
\end{axis}
\end{tikzpicture}%

%% file: Figures/age_stats_gender_0.tikz
% This file was created by matlab2tikz v0.4.6 running on MATLAB 8.3.
% Copyright (c) 2008--2014, Nico Schlömer <nico.schloemer@gmail.com>
% All rights reserved.
% Minimal pgfplots version: 1.3
% 
\begin{tikzpicture}

\begin{axis}[%
width=\tikzpicturewidth,
height=\tikzpictureheight,
axis on top,
font=\small,
clip=false,
scale only axis,
xmin=0.5,
xmax=5.5,
xtick={1,2,3,4,5},
xticklabels={\empty},
xlabel style={align=center},
xlabel={\\[1ex]Women being rated},
y dir=reverse,
ymin=0.5,
ymax=5.5,
ytick={1,2,3,4,5},
yticklabels={{18-21},{22-25},{26-29},{30-33},{34-37}},
ylabel={Men rating},
colorbar,
colormap/jet,
%colorbar style={font=\small, tick style={draw=none},xlabel={Attraction}, xlabel style={above=0.5mm}, yticklabel style={color=white}}
colorbar style={font=\small, ytick={0.1,0.9},yticklabels={min,max},tick style={draw opacity=0},width=10, xlabel={Attraction}, xlabel style={above=0.5mm}}
]
%colorbar style={title={},ytick={19,36},yticklabels={18,37},tick style={draw opacity=0},width=5},
%point meta min=18,
%point meta max=38

\addplot [forget plot] graphics [xmin=0.5,xmax=5.5,ymin=0.5,ymax=5.5] {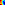};
\node[left, inner sep=0mm, rotate=90, text=black]
at (axis cs:0.98963133640553,5.529296875,1.77635683940025e-15) {18-21};
\node[left, inner sep=0mm, rotate=90, text=black]
at (axis cs:1.99193548387097,5.529296875,1.77635683940025e-15) {22-25};
\node[left, inner sep=0mm, rotate=90, text=black]
at (axis cs:2.99423963133641,5.529296875,1.77635683940025e-15) {26-29};
\node[left, inner sep=0mm, rotate=90, text=black]
at (axis cs:3.98502304147465,5.529296875,1.77635683940025e-15) {30-33};
\node[left, inner sep=0mm, rotate=90, text=black]
at (axis cs:4.98732718894009,5.529296875,1.77635683940025e-15) {34-37};
\end{axis}
\end{tikzpicture}%

%% file: Figures/hotness_paradox.tikz
% This file was created by matlab2tikz v0.4.6 running on MATLAB 8.3.
% Copyright (c) 2008--2014, Nico Schlömer <nico.schloemer@gmail.com>
% All rights reserved.
% Minimal pgfplots version: 1.3
% 
\begin{tikzpicture}

\begin{axis}[%
width=\tikzpicturewidth,
height=\tikzpictureheight,
scale only axis,
xmode=log,
xmin=1,
xmax=1000,
xminorticks=true,
xlabel={Number of neighbors},
ymin=20,
ymax=65,
ytick={20, 30, 40, 50, 60, 70},
ylabel={Hotter than you [\%]}
]
\addplot [color=blue,solid,line width=2.0pt,forget plot]
  table[row sep=crcr]{
1	46.3289081549716	\\
2	52.2758060146302	\\
3	55.2966675697643	\\
4	56.6242210782986	\\
5	57.7485776212409	\\
6	58.1278786236792	\\
7	59.1303169872663	\\
8	59.2928745597399	\\
9	59.8618260633975	\\
10	59.6179897046871	\\
20	61.1758331075589	\\
30	61.4061230018965	\\
40	61.7854240043349	\\
50	61.7854240043349	\\
60	61.7718775399621	\\
70	61.5144947168789	\\
80	61.9886209699269	\\
90	61.4874017881333	\\
100	61.5551341099973	\\
200	60.972636141967	\\
300	60.3088593876998	\\
400	59.9701977783798	\\
500	59.9701977783798	\\
600	60.0514765646166	\\
700	59.9701977783798	\\
800	60.0785694933622	\\
900	60.498509888919	\\
1000	60.5526957464102	\\
};
\node[fill=white, right, inner sep=0mm, text=blue, font=\footnotesize]
at (axis cs:31.6227766016838,61.4874017881333,0) {Men: visual features};
\addplot [color=blue,dashed,line width=2.0pt,forget plot]
  table[row sep=crcr]{
1	31.5361690598754	\\
2	29.9512327282579	\\
3	29.3822812246004	\\
4	28.6236792197237	\\
5	28.0276347873205	\\
6	26.9439176374966	\\
7	26.8219994581414	\\
8	26.4562449200759	\\
9	26.0633974532647	\\
10	25.9008398807911	\\
20	24.9119479815768	\\
30	24.3429964779193	\\
40	24.1262530479545	\\
50	23.7334055811433	\\
60	23.5979409374153	\\
70	23.4624762936874	\\
80	23.150907613113	\\
90	23.0696288268762	\\
100	22.9883500406394	\\
200	22.879978325657	\\
300	23.2050934706042	\\
400	23.7063126523977	\\
500	24.2210782985641	\\
600	24.5055540503929	\\
700	25.101598482796	\\
800	25.5350853427256	\\
900	26.0498509888919	\\
1000	26.6323489569222	\\
};
\node[fill=white, right, inner sep=0mm, text=blue, font=\footnotesize]
at (axis cs:31.6227766016838,24.1939853698185,0) {Men: Q};
\addplot [color=red,solid,line width=2.0pt,forget plot]
  table[row sep=crcr]{
1	48.989898989899	\\
2	51.9240019240019	\\
3	52.6214526214526	\\
4	52.4290524290524	\\
5	53.030303030303	\\
6	53.030303030303	\\
7	52.9341029341029	\\
8	52.8860028860029	\\
9	52.8379028379028	\\
10	52.4771524771525	\\
20	52.7417027417027	\\
30	52.6695526695527	\\
40	52.8138528138528	\\
50	53.1505531505532	\\
60	53.1746031746032	\\
70	53.1265031265031	\\
80	52.982202982203	\\
90	53.2708032708033	\\
100	53.030303030303	\\
200	53.2708032708033	\\
300	53.3910533910534	\\
400	53.4632034632035	\\
500	53.5113035113035	\\
600	53.5113035113035	\\
700	53.4632034632035	\\
800	53.2467532467533	\\
900	53.1986531986532	\\
1000	53.1505531505532	\\
};
\node[fill=white, right, inner sep=0mm, text=red, font=\footnotesize]
at (axis cs:31.6227766016838,52.7657527657528,0) {Women: visual features};
\addplot [color=red,dashed,line width=2.0pt,forget plot]
  table[row sep=crcr]{
1	43.0976430976431	\\
2	40.7166907166907	\\
3	39.6103896103896	\\
4	39.4179894179894	\\
5	39.0812890812891	\\
6	37.6382876382876	\\
7	37.1572871572872	\\
8	36.988936988937	\\
9	36.5079365079365	\\
10	35.4978354978355	\\
20	32.9004329004329	\\
30	31.4814814814815	\\
40	29.96632996633	\\
50	29.2688792688793	\\
60	29.028379028379	\\
70	28.3309283309283	\\
80	27.7537277537278	\\
90	27.5613275613276	\\
100	26.6474266474266	\\
200	24.4588744588745	\\
300	23.7614237614238	\\
400	22.5829725829726	\\
500	22.5589225589226	\\
600	22.7032227032227	\\
700	23.3285233285233	\\
800	23.977873977874	\\
900	24.7474747474748	\\
1000	25.5411255411255	\\
};
\node[fill=white, right, inner sep=0mm, text=red, font=\footnotesize]
at (axis cs:31.6227766016838,31.3852813852814,0) {Women: Q};
\addplot [color=black,dotted,line width=3.0pt,forget plot]
  table[row sep=crcr]{
1	50	\\
1000	50	\\
};
\node[fill=white, right, inner sep=0mm, text=black, font=\footnotesize]
at (axis cs:31.6227766016838,47,0) {Expected};
\end{axis}
\end{tikzpicture}%

%% file: Figures/nr_ratings_vs_acc.tikz
% This file was created by matlab2tikz v0.4.6 running on MATLAB 8.3.
% Copyright (c) 2008--2014, Nico Schlömer <nico.schloemer@gmail.com>
% All rights reserved.
% Minimal pgfplots version: 1.3
% 
\begin{tikzpicture}

\begin{axis}[%
width=\tikzpicturewidth,
height=\tikzpictureheight,
scale only axis,
font=\small,
xmode=log,
xmin=0.7,
xmax=1000,
xminorticks=true,
xlabel={Known ratings per user},
ymin=72.1635,
ymax=85,
ylabel={Accuracy [\%]},
legend style={at={(0.5,1.03)},anchor=north,xshift=-5mm,above=-5mm,draw=black,fill=white,legend cell align=left}
]
\addplot [color=black,dashed,line width=2.0pt]
  table[row sep=crcr]{
9	72.1635	\\
10	72.9187	\\
11	73.4895	\\
12	73.8725	\\
13	74.302	\\
14	74.6014	\\
15	74.8354	\\
16	75.1275	\\
17	75.2954	\\
18	75.4633	\\
19	75.6388	\\
20	75.8165	\\
25	76.5283	\\
30	76.9324	\\
35	77.2376	\\
40	77.4806	\\
45	77.7348	\\
50	77.9502	\\
60	78.464	\\
70	78.7808	\\
80	79.1411	\\
90	79.5008	\\
100	79.8209	\\
150	80.9712	\\
200	81.6311	\\
250	82.0272	\\
300	82.3663	\\
350	82.5566	\\
400	82.7299	\\
450	82.8656	\\
500	82.9764	\\
550	83.066	\\
600	83.1382	\\
650	83.1983	\\
700	83.2382	\\
750	83.2919	\\
800	83.3355	\\
850	83.366	\\
900	83.3914	\\
950	83.4121	\\
1000	83.4626	\\
};
\addlegendentry{Rating Query (MF)};

\addplot [color=black,solid,line width=2.0pt]
  table[row sep=crcr]{
0.7	75.3543	\\
1	75.8495	\\
2	76.305	\\
3	76.6971	\\
4	77.0665	\\
5	77.3847	\\
6	77.7747	\\
7	78.0432	\\
8	78.2744	\\
9	78.4812	\\
10	78.6772	\\
11	78.8693	\\
12	79.0144	\\
13	79.1622	\\
14	79.2883	\\
15	79.406	\\
16	79.518	\\
17	79.5838	\\
18	79.6741	\\
19	79.7675	\\
20	79.833	\\
25	80.1811	\\
30	80.4299	\\
35	80.6507	\\
40	80.8077	\\
45	80.9557	\\
50	81.0694	\\
60	81.3033	\\
70	81.4549	\\
80	81.6017	\\
90	81.7016	\\
100	81.8152	\\
150	82.1935	\\
200	82.4487	\\
250	82.6103	\\
300	82.7764	\\
350	82.8974	\\
400	82.9871	\\
450	83.0653	\\
500	83.127	\\
550	83.1768	\\
600	83.2375	\\
650	83.2822	\\
700	83.3321	\\
750	83.362	\\
800	83.4075	\\
850	83.4315	\\
900	83.4508	\\
950	83.4809	\\
1000	83.4922	\\
};
\addlegendentry{Rating Query (MF+VisReg)};

\addplot [color=red,solid,line width=2.0pt]
  table[row sep=crcr]{
0.7	75.91	\\
1000	75.91	\\
};
\addlegendentry{Visual Query (MF+VisReg)};

\end{axis}
\end{tikzpicture}%

%% file: Figures/hotness_q_gender_0.tikz
% This file was created by matlab2tikz v0.4.6 running on MATLAB 8.3.
% Copyright (c) 2008--2014, Nico Schlömer <nico.schloemer@gmail.com>
% All rights reserved.
% Minimal pgfplots version: 1.3
% 
\begin{tikzpicture}

\begin{axis}[%
width=\tikzpicturewidth,
height=\tikzpictureheight,
axis on top,
scale only axis,
xmin=0.5,
xmax=560.5,
y dir=reverse,
ymin=0.5,
ymax=420.5,
hide axis,
colormap/jet,
]
\addplot [forget plot] graphics [xmin=0.5,xmax=560.5,ymin=0.5,ymax=420.5] {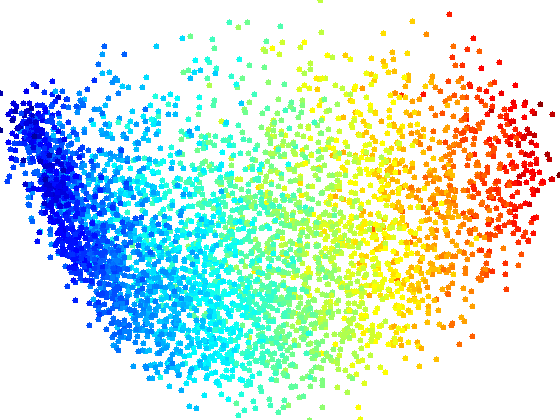};
\end{axis}
\end{tikzpicture}%

%% file: Figures/hotness_q_gender_1.tikz
% This file was created by matlab2tikz v0.4.6 running on MATLAB 8.3.
% Copyright (c) 2008--2014, Nico Schlömer <nico.schloemer@gmail.com>
% All rights reserved.
% Minimal pgfplots version: 1.3
% 
\begin{tikzpicture}

\begin{axis}[%
width=\tikzpicturewidth,
height=\tikzpictureheight,
axis on top,
scale only axis,
xmin=0.5,
xmax=560.5,
y dir=reverse,
ymin=0.5,
ymax=420.5,
hide axis,
colormap/jet,
colorbar,
colorbar style={title={},ytick={0.06,0.94},yticklabels={0\%,100\%},tick style={draw opacity=0},width=5},
point meta min=0,
point meta max=1
]
\addplot [forget plot] graphics [xmin=0.5,xmax=560.5,ymin=0.5,ymax=420.5] {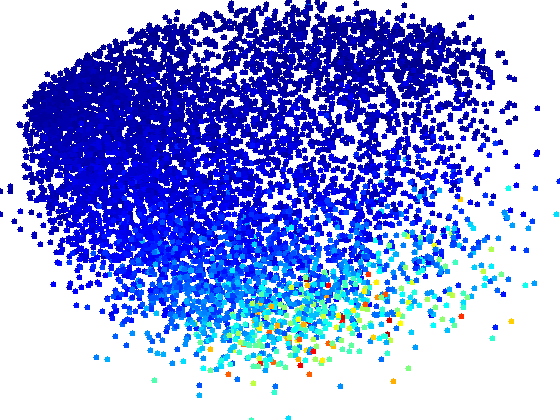};
\end{axis}
\end{tikzpicture}%

%% file: Figures/age_q_gender_0.tikz
% This file was created by matlab2tikz v0.4.6 running on MATLAB 8.3.
% Copyright (c) 2008--2014, Nico Schlömer <nico.schloemer@gmail.com>
% All rights reserved.
% Minimal pgfplots version: 1.3
% 
\begin{tikzpicture}

\begin{axis}[%
width=\tikzpicturewidth,
height=\tikzpictureheight,
axis on top,
scale only axis,
xmin=0.5,
xmax=560.5,
y dir=reverse,
ymin=0.5,
ymax=420.5,
hide axis,
colormap/jet,
]
\addplot [forget plot] graphics [xmin=0.5,xmax=560.5,ymin=0.5,ymax=420.5] {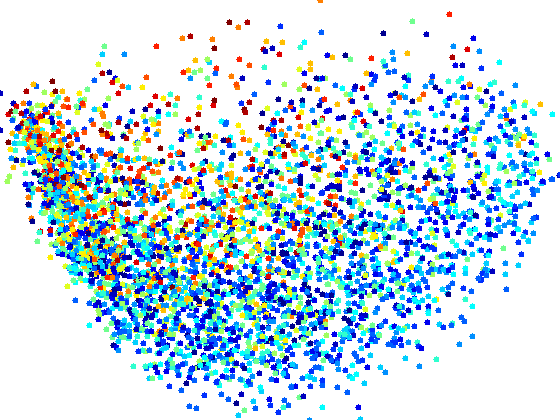};
\end{axis}
\end{tikzpicture}%

%% file: Figures/age_q_gender_1.tikz
% This file was created by matlab2tikz v0.4.6 running on MATLAB 8.3.
% Copyright (c) 2008--2014, Nico Schlömer <nico.schloemer@gmail.com>
% All rights reserved.
% Minimal pgfplots version: 1.3
% 
\begin{tikzpicture}

\begin{axis}[%
width=\tikzpicturewidth,
height=\tikzpictureheight,
axis on top,
scale only axis,
xmin=0.5,
xmax=560.5,
y dir=reverse,
ymin=0.5,
ymax=420.5,
hide axis,
colormap/jet,
colorbar,
colorbar style={title={},ytick={19,36},yticklabels={18,37},tick style={draw opacity=0},width=5},
point meta min=18,
point meta max=38
]
\addplot [forget plot] graphics [xmin=0.5,xmax=560.5,ymin=0.5,ymax=420.5] {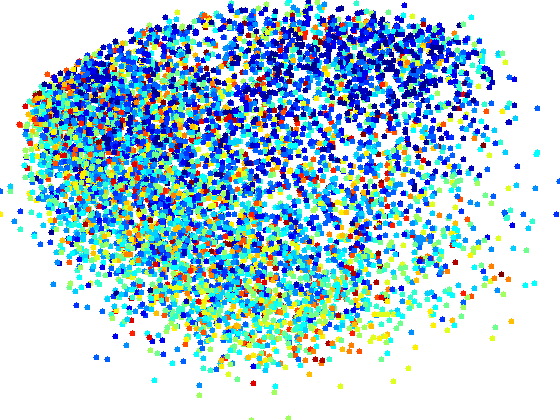};
\end{axis}
\end{tikzpicture}%

%% file: Figures/movielens_nr_ratings_vs_mae.tikz
% This file was created by matlab2tikz v0.4.6 running on MATLAB 8.3.
% Copyright (c) 2008--2014, Nico Schlömer <nico.schloemer@gmail.com>
% All rights reserved.
% Minimal pgfplots version: 1.3
% 
\begin{tikzpicture}

\begin{axis}[%
width=\tikzpicturewidth,
height=\tikzpictureheight,
scale only axis,
font=\small,
xmode=log,
xmin=0.7,
xmax=10000,
xminorticks=true,
xlabel={Known ratings per movie},
ymin=0.6,
ymax=1.1762,
ylabel={MAE},
legend style={at={(0.5,1.03)},anchor=north,xshift=17mm,draw=black,fill=white,legend cell align=left}
]
\addplot [color=black,dashed,line width=2.0pt]
  table[row sep=crcr]{
0.7	1.1762	\\
1	1.1612	\\
2	1.1466	\\
3	1.1317	\\
4	1.1169	\\
5	1.1021	\\
6	1.0872	\\
7	1.0733	\\
8	1.0593	\\
9	1.0452	\\
10	1.0311	\\
11	1.0178	\\
12	1.0042	\\
13	0.992	\\
14	0.9793	\\
15	0.9675	\\
16	0.9559	\\
17	0.9445	\\
18	0.9337	\\
19	0.9231	\\
20	0.9131	\\
25	0.8706	\\
30	0.8374	\\
35	0.8123	\\
40	0.7927	\\
45	0.7781	\\
50	0.7676	\\
60	0.754	\\
70	0.7468	\\
80	0.7432	\\
90	0.7413	\\
100	0.7399	\\
150	0.7366	\\
200	0.7343	\\
250	0.7319	\\
300	0.7297	\\
350	0.7278	\\
400	0.726	\\
450	0.7243	\\
500	0.7227	\\
550	0.7212	\\
600	0.7198	\\
650	0.7185	\\
700	0.7174	\\
750	0.7162	\\
800	0.7152	\\
850	0.7142	\\
900	0.7133	\\
950	0.7124	\\
1000	0.7116	\\
1100	0.7099	\\
1200	0.7086	\\
1300	0.7074	\\
1400	0.7063	\\
1500	0.7053	\\
1600	0.7045	\\
1700	0.7038	\\
1800	0.7031	\\
1900	0.7025	\\
2000	0.702	\\
2500	0.7	\\
3000	0.6989	\\
3500	0.698	\\
4000	0.6975	\\
4500	0.6972	\\
5000	0.6969	\\
5500	0.6968	\\
6000	0.6966	\\
6500	0.6965	\\
7000	0.6964	\\
7500	0.6963	\\
8000	0.6963	\\
8500	0.6962	\\
9000	0.6962	\\
9500	0.6961	\\
10000	0.6961	\\
};
\addlegendentry{Rating Query (MF)};

\addplot [color=black,solid,line width=2.0pt]
  table[row sep=crcr]{
0.7	0.9165	\\
1	0.9113	\\
2	0.9056	\\
3	0.9006	\\
4	0.897	\\
5	0.8919	\\
6	0.8875	\\
7	0.8831	\\
8	0.8791	\\
9	0.8755	\\
10	0.8724	\\
11	0.8693	\\
12	0.8666	\\
13	0.8634	\\
14	0.8607	\\
15	0.8586	\\
16	0.8564	\\
17	0.8536	\\
18	0.8512	\\
19	0.8493	\\
20	0.848	\\
25	0.8396	\\
30	0.832	\\
35	0.8265	\\
40	0.8219	\\
45	0.8174	\\
50	0.8118	\\
60	0.8036	\\
70	0.797	\\
80	0.7911	\\
90	0.7857	\\
100	0.7803	\\
150	0.7608	\\
200	0.75	\\
250	0.743	\\
300	0.738	\\
350	0.7343	\\
400	0.7312	\\
450	0.7287	\\
500	0.7265	\\
550	0.7245	\\
600	0.7227	\\
650	0.7211	\\
700	0.7197	\\
750	0.7183	\\
800	0.7171	\\
850	0.7161	\\
900	0.715	\\
950	0.714	\\
1000	0.713	\\
1100	0.7112	\\
1200	0.7098	\\
1300	0.7085	\\
1400	0.7072	\\
1500	0.7062	\\
1600	0.7053	\\
1700	0.7045	\\
1800	0.7038	\\
1900	0.7032	\\
2000	0.7026	\\
2500	0.7005	\\
3000	0.6992	\\
3500	0.6983	\\
4000	0.6978	\\
4500	0.6974	\\
5000	0.6972	\\
5500	0.697	\\
6000	0.6968	\\
6500	0.6967	\\
7000	0.6966	\\
7500	0.6965	\\
8000	0.6965	\\
8500	0.6964	\\
9000	0.6964	\\
9500	0.6964	\\
10000	0.6963	\\
};
\addlegendentry{Rating Query (MF+VisReg)};

\addplot [color=red,solid,line width=2.0pt]
  table[row sep=crcr]{
0.7	0.813	\\
10000	0.813	\\
};
\addlegendentry{Visual Query (MF+VisReg)};

\end{axis}
\end{tikzpicture}%